\documentclass[letterpaper, 10 pt, conference]{ieeeconf}  

\IEEEoverridecommandlockouts                              

\usepackage{amsmath}
\usepackage{amssymb}
\usepackage{graphicx}
\usepackage{textcomp}
\usepackage{xcolor}
\usepackage{pgfplots}
\usepackage{subfigure}
\usepgfplotslibrary{groupplots}
\usepackage{tikz}
\usepackage{tikzscale}
\usepackage{hyperref}
\usepackage{cleveref}
\usepackage{float}
\usepackage{multirow}
\usepackage{caption}
\usepackage{algorithm}
\usepackage{algpseudocode} 
\usepackage{siunitx}
\usepackage{comment}
\usepackage{booktabs}
\usepackage{alphalph}

\crefname{figure}{Fig.}{Fig.}
\Crefname{figure}{Figure}{Figures}
\crefname{equation}{}{}
\Crefname{equation}{Equation}{Equations}

\newcommand{\ie}{\textit{i}.\textit{e}., }
\newcommand{\eg}{\textit{e}.\textit{g}., }

\DeclareMathOperator*{\argmax}{arg\,max}

\title{\LARGE \bf
Simulation-aided Learning from Demonstration for \\ Robotic LEGO Construction
}

\author{Ruixuan Liu$^{1}$, Alan Chen$^{2}$, Xusheng Luo$^{1}$ and Changliu Liu$^{1}$ 
\thanks{*This work is in part supported by Siemens and Manufacturing Futures Institute, Carnegie Mellon University, through a grant from the Richard King Mellon Foundation.}
\thanks{$^{1}$Ruixuan Liu, Xusheng Luo, and Changliu Liu are with Robotics Institute,
	Carnegie Mellon University,
	Pittsburgh, PA, 15213, USA.
        {\tt\small ruixuanl, xushengl, cliu6@andrew.cmu.edu}%
}
\thanks{$^{2}$Alan Chen is with Westlake Highschool, 
        Austin, TX, 78733, USA. 
        {\tt\small alanjiach@gmail.com}
}}

\begin{document}

\maketitle
\thispagestyle{empty}
\pagestyle{empty}

\begin{abstract}
Recent advancements in manufacturing have a growing demand for fast, automatic prototyping (\ie assembly and disassembly) capabilities to meet users’ needs. 
This paper studies automatic rapid LEGO prototyping, which is devoted to constructing target LEGO objects that satisfy individual customization needs and allow users to freely construct their novel designs.
A construction plan is needed in order to automatically construct the user-specified LEGO design.
However, a freely designed LEGO object might not have an existing construction plan, and generating such a LEGO construction plan requires a non-trivial effort since it requires accounting for numerous constraints (\eg object shape, colors, stability, etc.).
In addition, programming the prototyping skill for the robot requires the users to have expert programming skills, which makes the task beyond the reach of the general public.
To address the challenges, this paper presents a simulation-aided learning from demonstration (SaLfD) framework for easily deploying LEGO prototyping capability to robots. 
In particular, the user demonstrates constructing the customized novel LEGO object. 
The robot extracts the task information by observing the human operation and generates the construction plan.
A simulation is developed to verify the correctness of the learned construction plan and the resulting LEGO prototype.
The proposed system is deployed to a FANUC LR-mate 200id/7L robot.
Experiments demonstrate that the proposed SaLfD framework can effectively correct and learn the prototyping (\ie assembly and disassembly) tasks from human demonstrations.
And the learned prototyping tasks are realized by the FANUC robot.

\end{abstract}

\section{Introduction}

Recent advancements in manufacturing have a growing demand for fast, automatic prototyping capabilities for assembly tasks \cite{10.5555/3113190.3113359,AHMAD2015412,GOMESDESA1999389} to meet users’ needs.
LEGO is a well-known platform for assembly prototyping and constructing proof-of-concepts \cite{ZHOU2020103282}.
There are a wide variety of LEGO bricks with different shapes and colors, which allows individual customization.
This paper studies automatic rapid LEGO prototyping, which is devoted to constructing target LEGO objects that satisfy individual needs and is easy to adopt by end users.
Recent work \cite{liu2023lightweight} enables robots to robustly manipulate LEGO bricks.
With the capability of manipulating LEGO bricks, the robot can construct the target LEGO object given a construction plan.

\begin{figure}
    \centering
    \includegraphics[width=\linewidth]{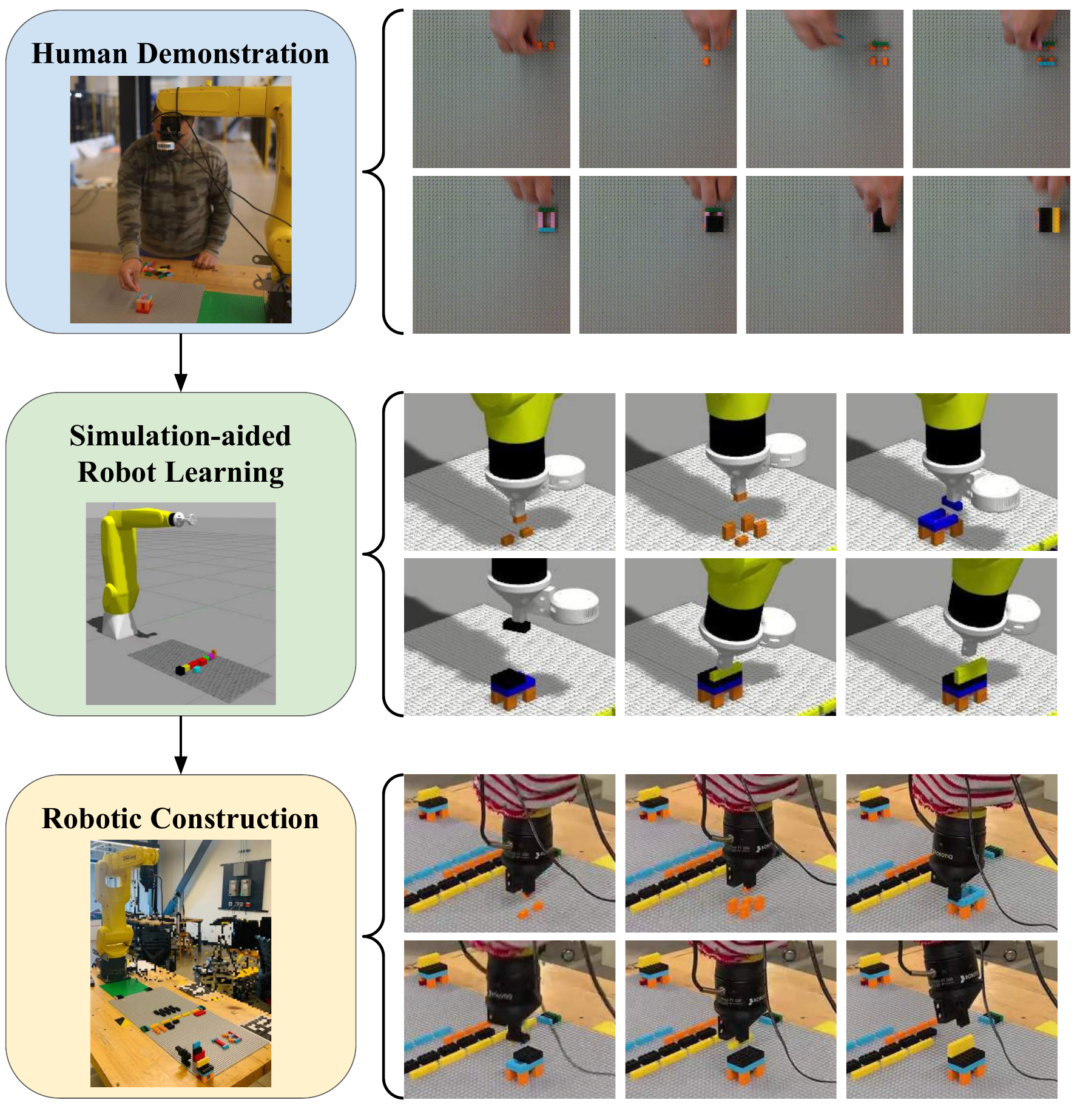}
    \vspace{-20pt}
    \caption{\footnotesize Illustration of the pipeline of simulation-aided learning from demonstration for robotic LEGO construction. \label{fig:lfd_pipeline}}
    \vspace{-20pt}
\end{figure}

\begin{figure}
\subfigure[Brick usage.]{\includegraphics[width=0.32\linewidth]{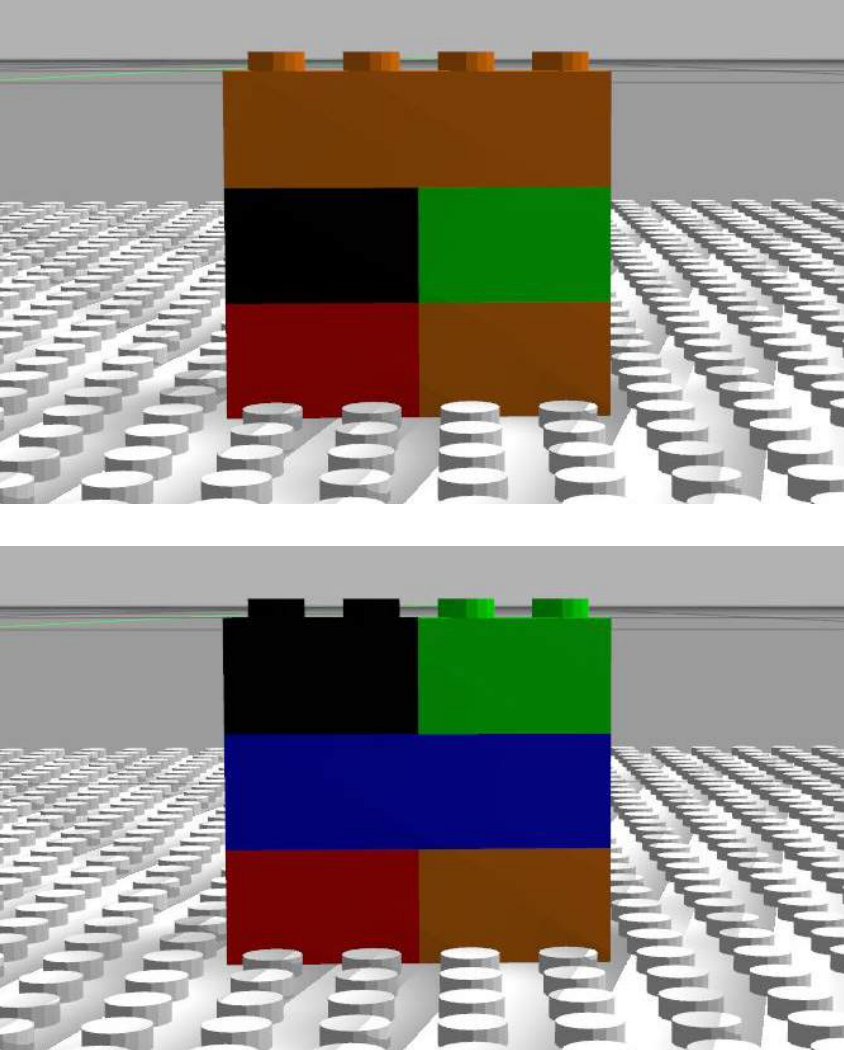}\label{fig:brick_usage}}\hfill
\subfigure[Appearance.]{\includegraphics[width=0.32\linewidth]{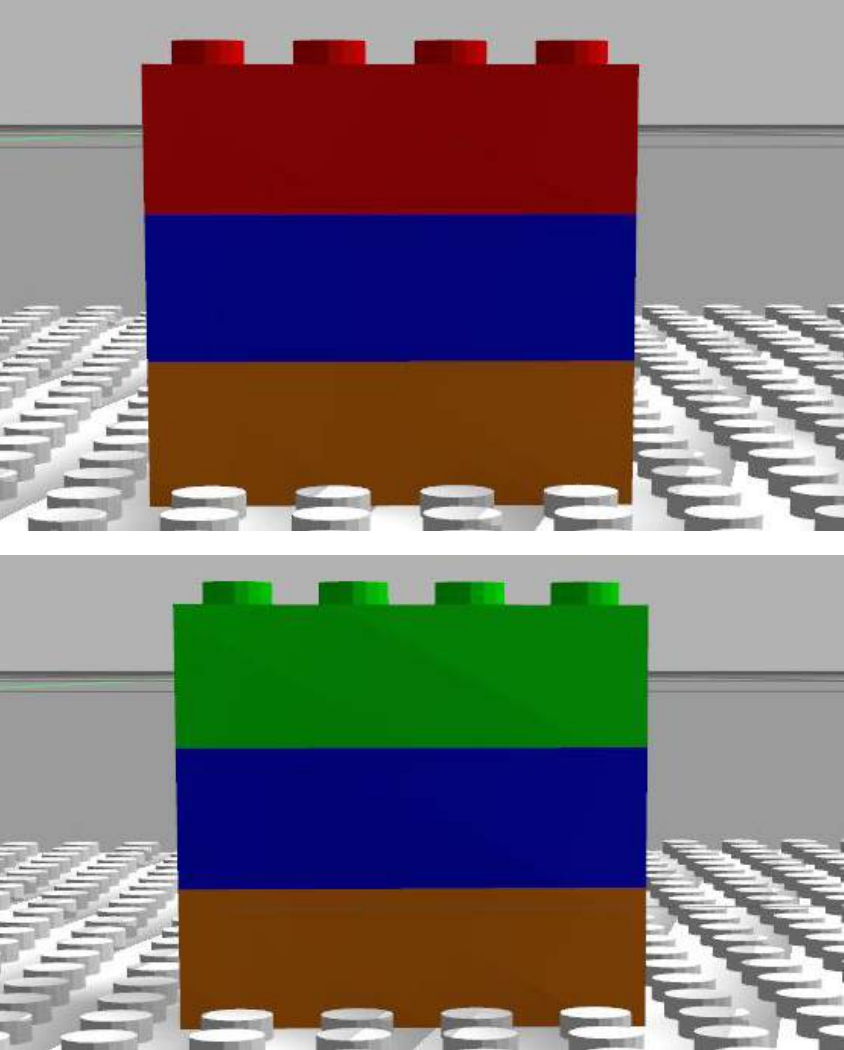}\label{fig:color}}\hfill
\subfigure[Stability.]{\includegraphics[width=0.32\linewidth]{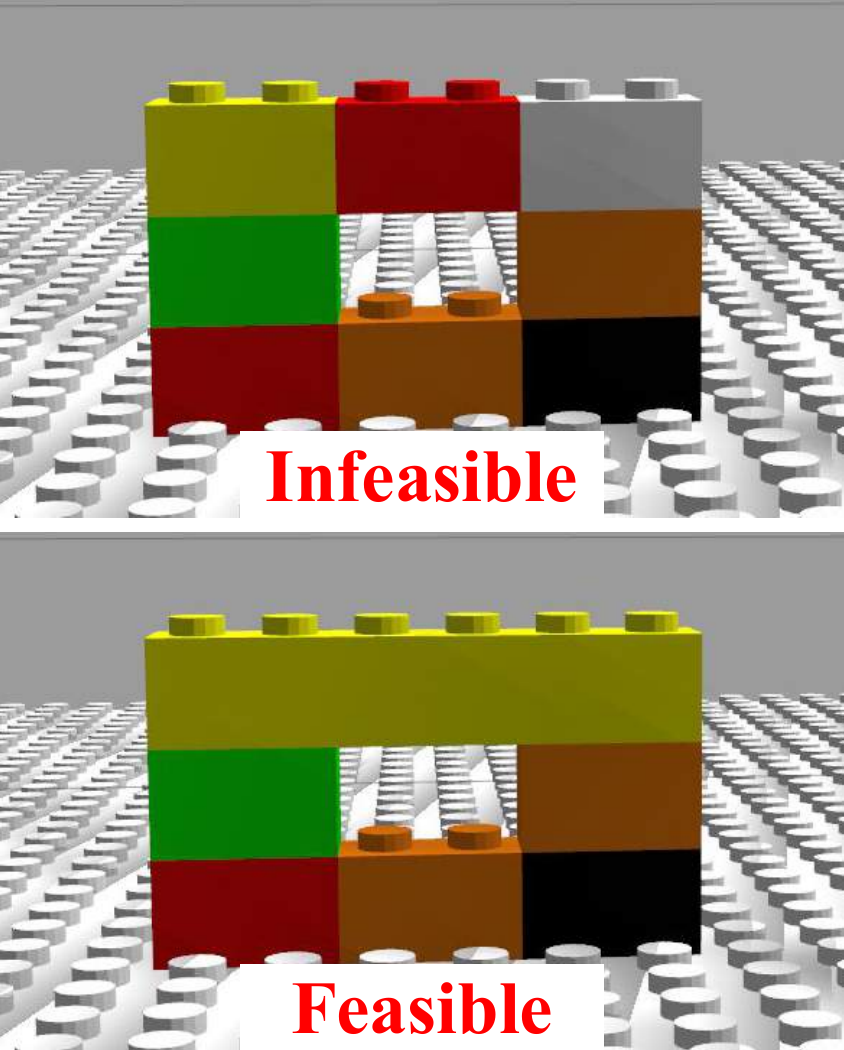}\label{fig:geometric}}\hfill
\vspace{-5pt}
    \caption{\footnotesize Constraints in LEGO construction plans. \label{fig:constraints}}
    \vspace{-20pt}
\end{figure}

However, a construction plan is usually difficult to generate, even in the case that the shape of the LEGO object is given, since numerous constraints need to be considered in order to satisfy individual needs.
\Cref{fig:constraints} illustrates constraints that need to be considered when generating a LEGO construction plan.
First, the construction plan should accommodate individual preferences on brick usage. 
\Cref{fig:brick_usage} displays two identical structures. However, the top one has a 1x4 brick on the top layer, whereas the bottom one has two 1x2 bricks on the top layer.
Second, the construction plan should account for individual preferences on object appearances, \eg colors. 
\Cref{fig:color} shows two identical structures with different appearances due to using bricks with different colors.
Third, the construction plan should generate geometrically feasible objects.
\Cref{fig:geometric} illustrates two identical objects with different brick arrangements. 
The top one is geometrically infeasible since it has a floating brick on the top layer.
Whereas the bottom one is stable and feasible to be built.
Besides the above-mentioned constraints, generating a machine-readable construction plan is time-consuming and requires expert knowledge, which would hinder the deployment to the general public.
It is critical that the construction plan is feasible by considering the above-mentioned requirements as well as easy to generate so that the robot can successfully realize the LEGO prototypes and the system can benefit more people.

To address the challenges, this paper studies generating LEGO construction plans by learning from human demonstrations (LfD) and proposes a pipeline for simulation-aided learning from demonstration (SaLfD) as shown in \cref{fig:lfd_pipeline}.
The user first demonstrates building the intended LEGO object.
The robot learns the assembly and disassembly tasks from the demonstration and replicates the LEGO construction.
In particular, this paper presents a closed-loop learning framework to correct potentially infeasible LEGO construction plans.
The framework consists of two major modules: 1) task extraction from demonstration, and 2) simulation verification.
The first module captures human operation and extracts the assembly task sequence.
Since the human demonstrates assembly in the real world, the learned construction plan would inherently satisfy the constraints in \cref{fig:brick_usage,fig:color,fig:geometric}.
To ensure the correctness of the learned construction plan, a simulation is developed to assist in verifying the learned task information and refine the construction plan.
The proposed SaLfD framework is deployed to a FANUC LR-mate 200id/7L robot. 
The experiments demonstrate that our system can effectively correct and learn the assembly and disassembly tasks from intuitive human demonstrations.
The learned tasks are realized by a FANUC robot arm.

\section{Related Works}
\subsection{LEGO Construction}
LEGO construction has been recently widely studied \cite{Kim2014SurveyOA}.
Early works \cite{gower1998lego,smal2008automated} use search algorithms to generate LEGO structures.
\cite{petrovic2001solving,peysakhov2003using} formulate LEGO construction as an optimization problem and use evolutionary algorithms to optimize the LEGO structures.
\cite{LEGOBuilder,testuz2013automatic} voxelize the 3D model of the target object and initialize the voxel with unit 1x1 LEGO bricks. 
The LEGO structure is solved using greedy algorithms by rearranging bricks.
\cite{testuz2013automatic,10.1145/2816795.2818091,8419684,KOLLSKER2021270} consider the structure stability and generate physically-realizable LEGO structures.
\cite{doi:10.1177/09544054211053616,10.1111:cgf.13603} preserve the visual details of the input 3D model and generate a complex LEGO structure.
Recent works \cite{thompson2020LEGO,lennon2021image2lego} use learning-based techniques to generate LEGO structures.
\cite{ChungH2021neurips,KimJ2020arxiv,ahn2022sequential} study the sequential assembly problem and generate LEGO structures to fill up the target 3D model.
Given a LEGO model, \cite{10.1007/978-3-031-19815-1_6} leverages a simulator to understand the LEGO structure and the relationships between bricks.
The underlying assumption of existing works when generating the construction plan is that the 3D model (either mesh or voxel) of the target object is known.
However, there are a limited number of 3D models of commonsense objects available.
In fact, when people are creating a new object for a specific use, the 3D model is usually unknown and it would be difficult for the general public to obtain the 3D model before building it.
It is possible to first define an instruction and then translate it to machine interpretable LEGO construction plan \cite{wang2022translating}.
However, defining such an instruction is time-consuming.

\subsection{Learning from Demonstration}

\begin{figure}
\subfigure[Color ambiguity: a black 2x4 brick is placed in the red bounding box.]{\includegraphics[width=0.49\linewidth]{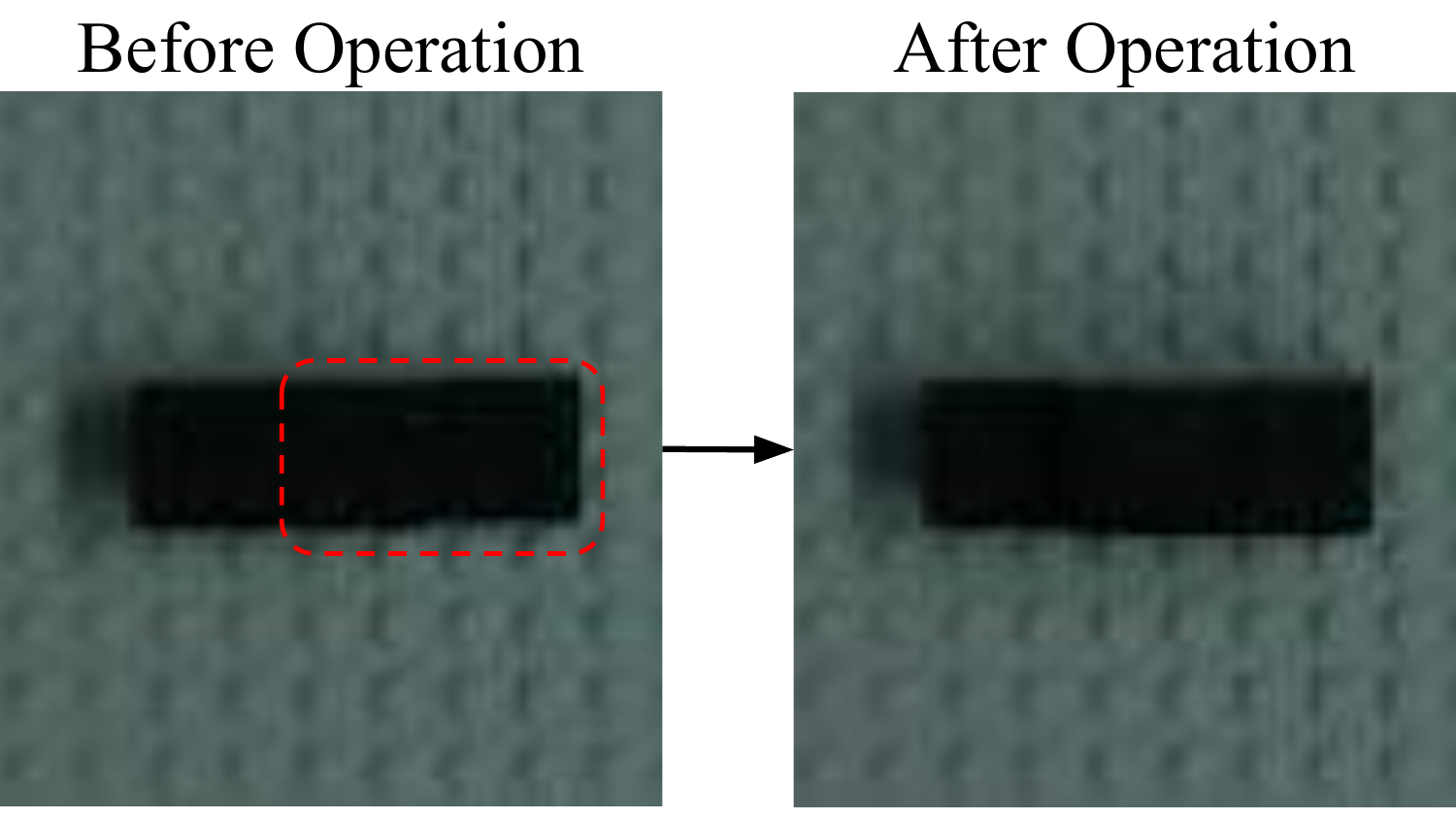}\label{fig:color_ambiguity}}
\subfigure[Depth ambiguity: two identical bricks gives different depth measurements.]{\includegraphics[width=0.49\linewidth]{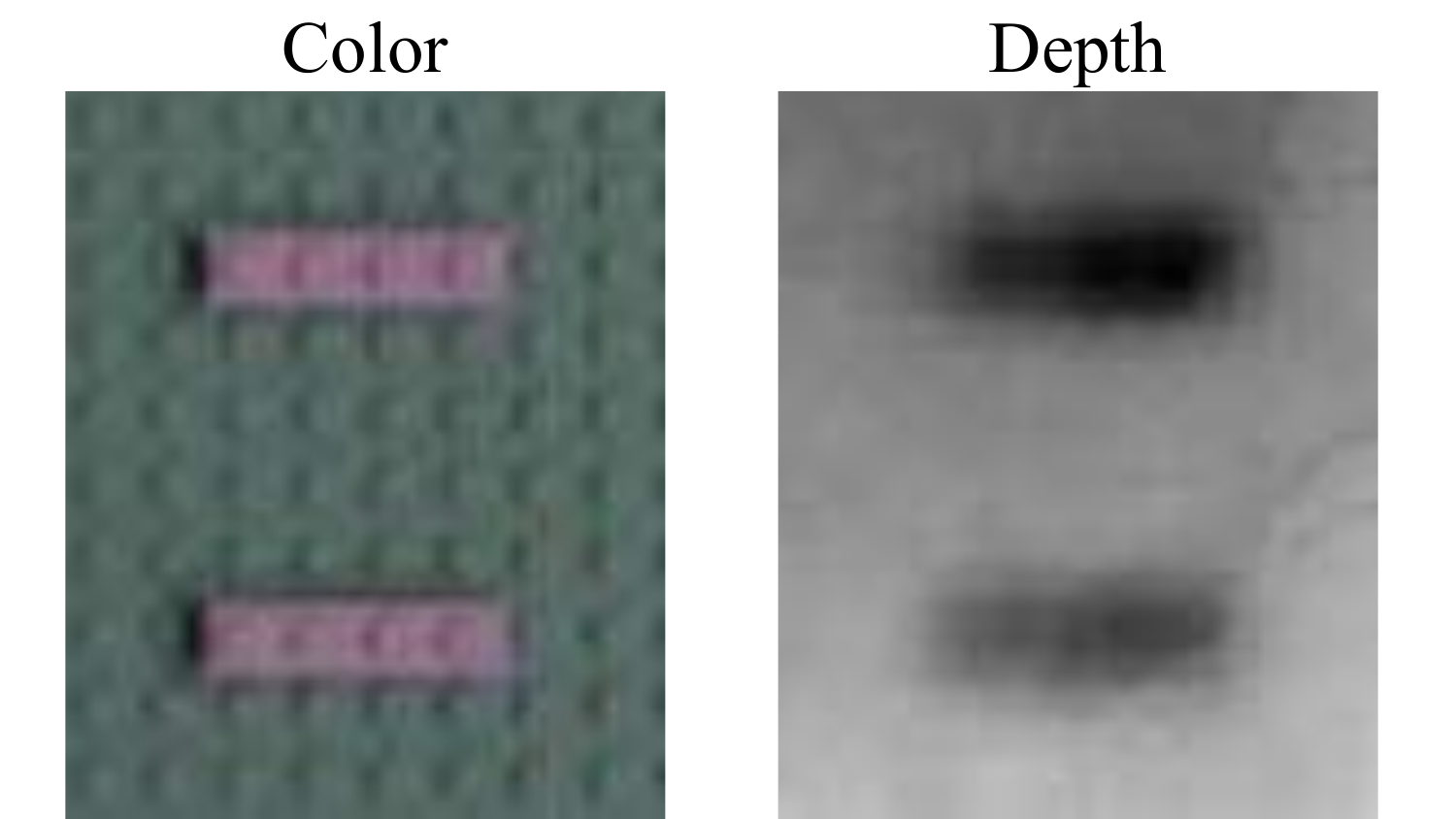}\label{fig:depth_ambiguity}}
\vspace{-5pt}
    \caption{\footnotesize Challenges in LEGO LfD. \label{fig:Challenges}}
    \vspace{-20pt}
\end{figure}

LfD has been widely studied \cite{doi:10.1146/annurev-control-100819-063206,10.1016/j.robot.2008.10.024,9074946}.
Many works focus on learning the robot control policy (\eg imitation learning) \cite{7989334, 10161416}. 
On the other hand, programming by demonstration (PbD) \cite{robotics7020017} is another category, where people develop easy approaches to program robots and lower the requirements for the general public to use robots. 
\cite{7451754} compares different PbD approaches, including kinesthetic teaching, teleoperating, etc.
\cite{Stenmark2016FromDT,7989334} teach the robot to manipulate LEGO bricks by kinesthetic teaching.
\cite{7606936} learns the assembly task from visual human demonstrations and refines the learned task using a scene parser in a probabilistic manner.
The human demonstration is captured using a multi-camera system, which might suffer from color ambiguity due to camera viewpoints.
To lower the cost, recent work \cite{liu2023robotic} facilitates an onboard wrist camera (shown in \cref{fig:setup}) to capture and learn the 2D LEGO construction from human demonstration. 
However, LfD for general 3D LEGO construction remains challenging for several reasons.
First, the color ambiguity would confuse the visual camera.
\Cref{fig:color_ambiguity} illustrates the color ambiguity. 
A black 2x4 brick is assembled at the place indicated by the red bounding box, but such operation is hard to capture by regular RGB sensors since the assembled brick blends into the background.
To address the challenge, we use an RGB-D camera to capture the human demonstration, in which both color and depth information can be captured.
The single-camera setup keeps the cost low and is easy to set up.
Second, due to the tiny sizes and reflective surfaces, the depth measurements of LEGO bricks are noisy and inconsistent.
\Cref{fig:depth_ambiguity} illustrates the depth ambiguity.
Given two identical bricks, the depth measurements for the two bricks can be very different, which significantly increases the problem complexity.

\section{Problem Formulation}

\begin{figure*}
    \centering
    \includegraphics[width=\linewidth]{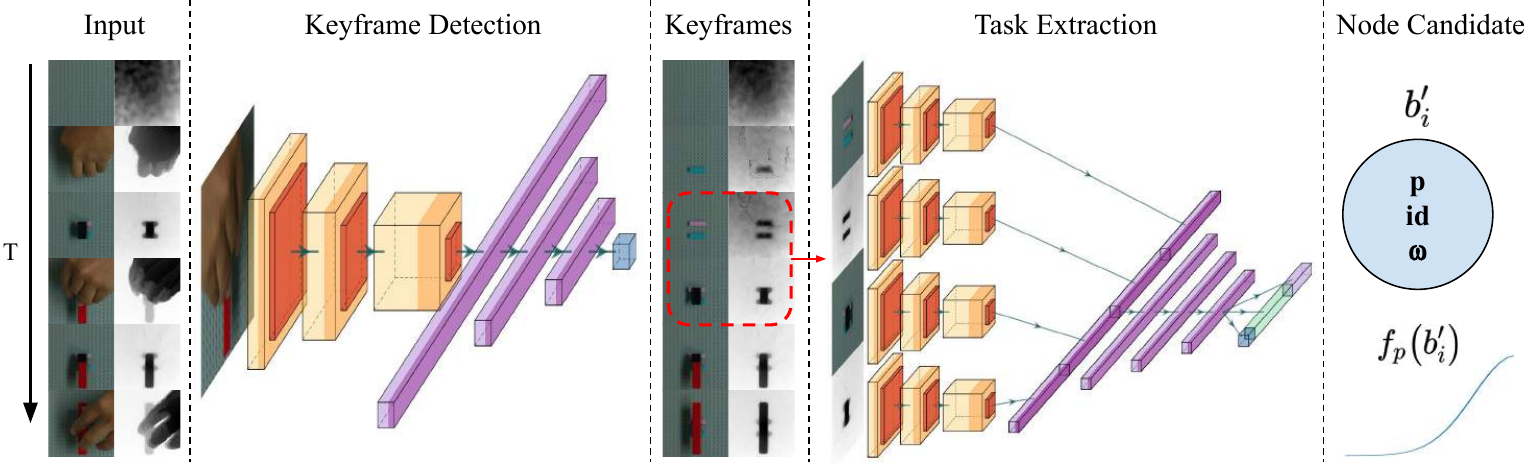}
    \vspace{-5pt}
    \caption{\footnotesize Illustration of the task extraction from human demonstration. \label{fig:pipeline}}
    \vspace{-20pt}
\end{figure*}

This paper considers the scenario where the user is creating novel objects on a LEGO baseplate (shown in \cref{fig:setup}), and thus, the target object model is unknown.
A LEGO structure can be represented as a set of bricks $B=\{b_1;~b_2;\dots;~b_n\}$, where $n$ is the number of bricks. And $b_i=[p;~id;~\omega]$, which encodes the brick position, brick type, and orientation.
The goal is to generate a structure $B$ that is identical to the target structure $B^t$.
The problem can be formulated as
\begin{equation}\label{eq:prob}
    \min L(B^t, B),
\end{equation}
where $L(\cdot)$ is a cost function calculating the difference between two LEGO structures. 
If the target object model is known, existing works \cite{10.1145/2816795.2818091,KOLLSKER2021270} can solve $B$.
However, the target model is unknown when users are creating novel LEGO designs. 
To address the challenges (\ie color and depth ambiguities in \cref{fig:Challenges}, unknown target model, do not require expert knowledge), we propose a SaLfD pipeline to solve \cref{eq:prob}.
This paper focuses on learning the assembly task information for LEGO construction, whereas the robot's skill of manipulating LEGO bricks is addressed in \cite{liu2023lightweight}.

\section{Simulation-aided Learning from Human Demonstration}

The SaLfD pipeline consists of two major components: 1) task extraction from demonstration, where the assembly task $b_i$ is learned, and 2) simulation verification, where $b_i$ is refined for correct assembly.

\subsection{Task Extraction from Demonstration}

LEGO is constructed brick by brick, and therefore, \cref{eq:prob} can be written as 
\begin{equation}\label{eq:minPi}
\begin{split}
    \min_{b_i} L(b_i, b^t_i)=L_p(p_i, p^t_i) + L_i(id_i, id^t_i)+L_w(\omega_i, \omega^t_i),
    \end{split}
\end{equation}
in which we decouple \cref{eq:prob} to an optimization problem for each brick operation.

\paragraph{Keyframe Detection}
Given an image sequence of the demonstration, only frames that display the state of the object matter, whereas the frames that the human is operating are not useful.
Therefore, it is desired to filter out the unhelpful frames and only keep the frames showing the start and end states of each brick operation.

Our pipeline uses a convolution neural network (CNN) \cite{726791} to detect the keyframes.
The architecture is shown in \cref{fig:pipeline}.
The network has 3 convolution layers, each followed by a ReLU activation layer and a max-pooling layer.
It then passes through 3 fully connected layers with the ReLU activation function.
The module takes an RGB image as input and outputs a binary label indicating whether it is a keyframe.
With a sequence of classified binary keyframe labels, a sliding filter is applied to extract a sequence of keyframes as shown in the middle column in \cref{fig:pipeline}.
Essentially, the network detects the human presence in the scene.
The sliding filter selects the keyframe with the highest confidence among a sequence of keyframes.
The resulting keyframe sequence indicates the start and end states of each human operation.

\paragraph{Task Extraction}

Given the sequence of the keyframes, the goal is to extract the information $b_i$ for each brick operation.
\cite{liu2023robotic} uses a rule-based approach by subtracting two consecutive keyframes and obtaining the feature mask. 
The task information $b_i$ is extracted from the feature mask. 
However, due to the noisy depth measurement, it is difficult to explicitly design a rule for extracting the feature for both color and depth inputs.

Therefore, we use data-driven approaches to solve \cref{eq:minPi}. 
In particular, a CNN is adopted to build the task extraction module.
The input is a pair of consecutive RGB-D keyframes as shown in the red bounding box in \cref{fig:pipeline}.
Due to the uncertainty shown in \cref{fig:Challenges}, the module outputs a ranked list of node candidates based on confidence.
The right side of \cref{fig:pipeline} illustrates the architecture of the task extraction network.
A sliding window selects the consecutive RGB-D keyframes.
The network has an individual channel with 3 convolution layers for each image input. 
After that, the encoded hidden features are concatenated and passed through 4 fully connected layers with the ReLU activation function.
The network outputs the brick ID label, the orientation label, and the position. 
Determining the brick ID is a classification problem. 
Essentially, the module classifies the brick being assembled into one of the $N$ possible LEGO bricks available, \ie $id\in [1,2,\dots,N]$.
Similarly, $\omega\in[0,1]$ is a binary feature due to the nature of LEGO assembly. 
A brick can only be placed horizontally or vertically.
Determining the brick position $p=[x,y,z]$ is a regression problem.
Since LEGO construction discretizes the world, the brick position is determined in a discrete space, \ie $x\in[1,\dots X]$, $y\in[1,\dots Y]$, $z\in[1,\dots Z]$, where $X,Y,Z$ are the boundary of the construction space.
Since uncertainty exists in the RGB-D images, we want the module output to be a probabilistic $p$ instead of deterministic.
However, it is difficult to generate a probability distribution of $p$ in the entire 3D workspace.
Therefore, we assume the uncertainty follows a Gaussian distribution, and the task extraction network outputs the mean values of the distribution, \ie $\mu_{p}=[\mu_x, \mu_y,\mu_z]$.
The position distribution is then calculated as 
\begin{equation}\label{eq:multi_normal}
\begin{split}
    f_p(p,\mu_{p})&=\frac{1}{(2\pi)^{\frac{3}{2}}|\Sigma|^{\frac{1}{2}}}\exp{-\frac{1}{2}\delta_{p}^T\Sigma^{-1}\delta_{p}},\\
    \delta_{p} &= p-\mu_{p},
    \end{split}
\end{equation}
where $\Sigma$ is the covariance matrix, which is determined by $||\mu_{p} - r(\mu_{p})||$.
And $r(\cdot)$ rounds $\mu_{p}$ to the nearest integers.
The confidence of the node candidate is calculated as $f(b_i')=f_p(p,\mu_{p})\cdot f_{id} \cdot f_{\omega}$, where $f_{id}$ and $f_{\omega}$ are the classification confidence.

\subsection{Simulation Verification}

\begin{figure}
\subfigure[Digital twin]{\includegraphics[width=0.45\linewidth]{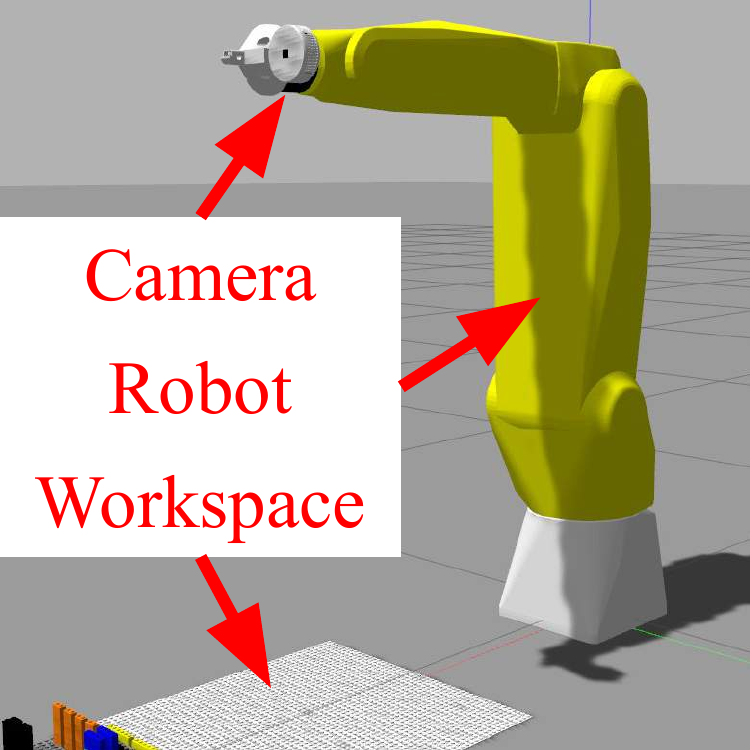}\label{fig:digital_twin}}\hfill
\subfigure[Real setup.]{\includegraphics[width=0.45\linewidth]{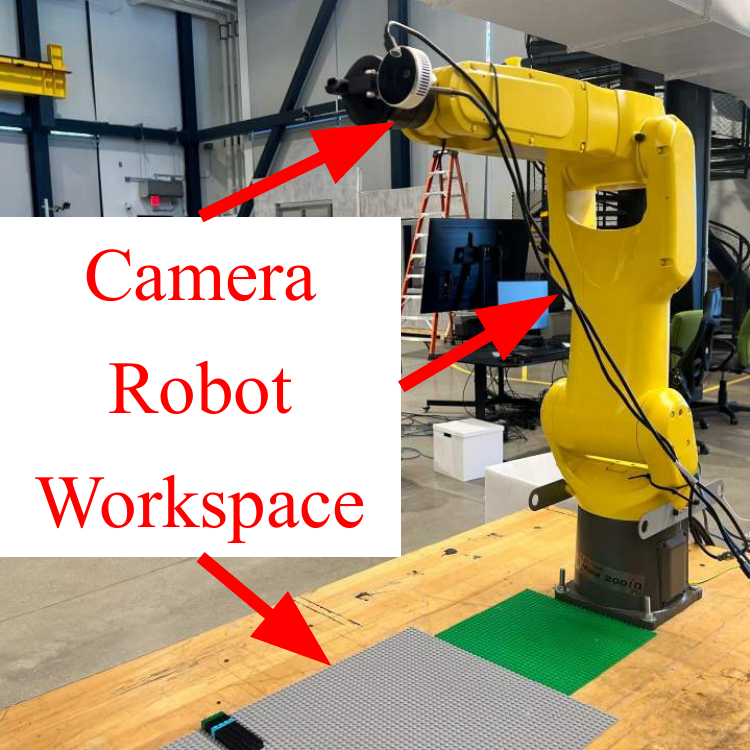}\label{fig:setup}}\hfill
\subfigure[Pipeline of the simulation verification.]{\includegraphics[width=\linewidth]{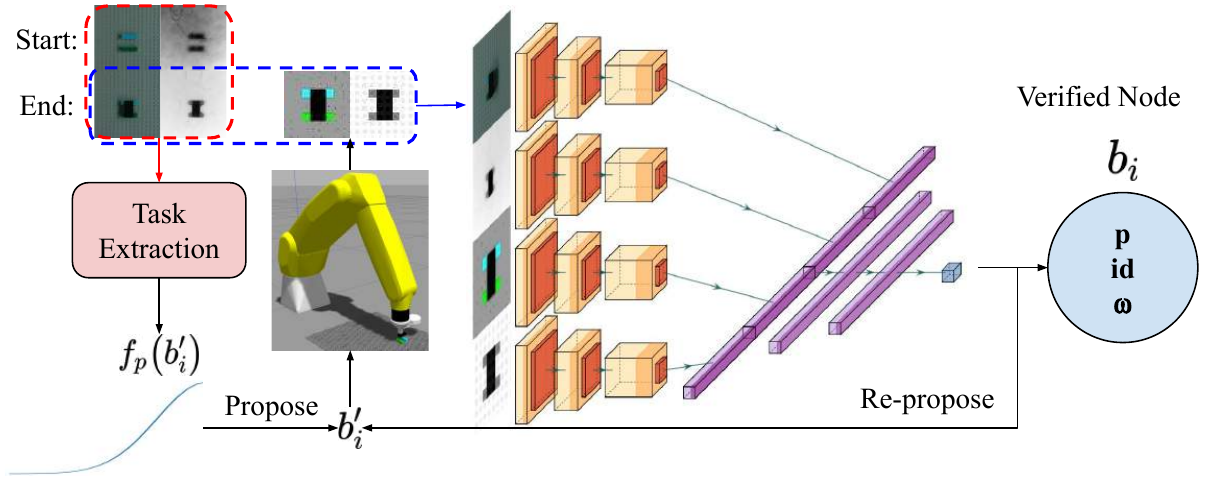}\label{fig:digital_twin_verification}}
\vspace{-5pt}
    \caption{\footnotesize Illustration of the simulation verification. \label{fig:dg_verification}}
    \vspace{-20pt}
\end{figure}

Due to the uncertainty in \cref{fig:Challenges}, it is difficult to ensure that the highest ranked $b_i'$ is indeed correct.
Therefore, we use simulation verification to alleviate the noisy measurement and verify the correctness of $b_i'$.
The intuition is that the simulation offers controllable clean measurements, which can be used as a reference.
\Cref{fig:digital_twin} illustrates the simulation environment, which uses the identical robot as in the real world, and shares the same robot control with the real system. 
It also mimics the environment setup (\ie the workspace and LEGO bricks) as shown in \cref{fig:digital_twin,fig:setup}.
The simulation is developed in ROS Gazebo \cite{1389727}.
Due to the limitation of the camera model in Gazebo, the camera field-of-view in the simulation is slightly different from the real camera.

\Cref{fig:digital_twin_verification} illustrates the details of the simulation verification.
In particular, given a ranked list of node candidates $b_i'$ from $f(b_i')$, the simulation executes the highest ranked $b_i'$ and compares the resulting LEGO state with the observations (\ie color and depth) in the real environment.
A similar CNN, with 4 channels of 3 convolution layers and 3 fully connected layers, is used to output a confidence score $s$ indicating the similarity between the end states in the real environment and the simulation.
If $s > \delta_s$ (a user-defined confidence threshold), then $b_i$ is verified.
Otherwise, the $b_i'$ with the second highest confidence is then proposed for re-verification. 
The verification continues until a confident $b_i'$ is found.
If no candidate satisfies $s > \delta_s$, then the final node is determined by solving $\argmax_{p, id, \omega}s$.

\section{Experiments}
To demonstrate our SaLfD system, we deploy it to a FANUC LR-mate 200id/7L robot as shown in \cref{fig:setup}.
A Realsense L515 is integrated into the EOAT \cite{liu2023lightweight} via the extension slots for capturing the human demonstration.
Two 48x48 LEGO plates are placed in front of the robot.
One is for brick storage (\ie a pre-defined setup) and the other one is for construction.
The user demonstrates the assembly on the construction plate.
The robot uses the bricks on the storage plate and builds on the construction plate.
The keyframe detection and task extraction are trained using demonstration data in the real setup.
The simulation verification model is trained using data both in the real setup and the simulation.

\begin{figure}
\centering
\subfigure[AI.]{\includegraphics[width=0.24\linewidth]{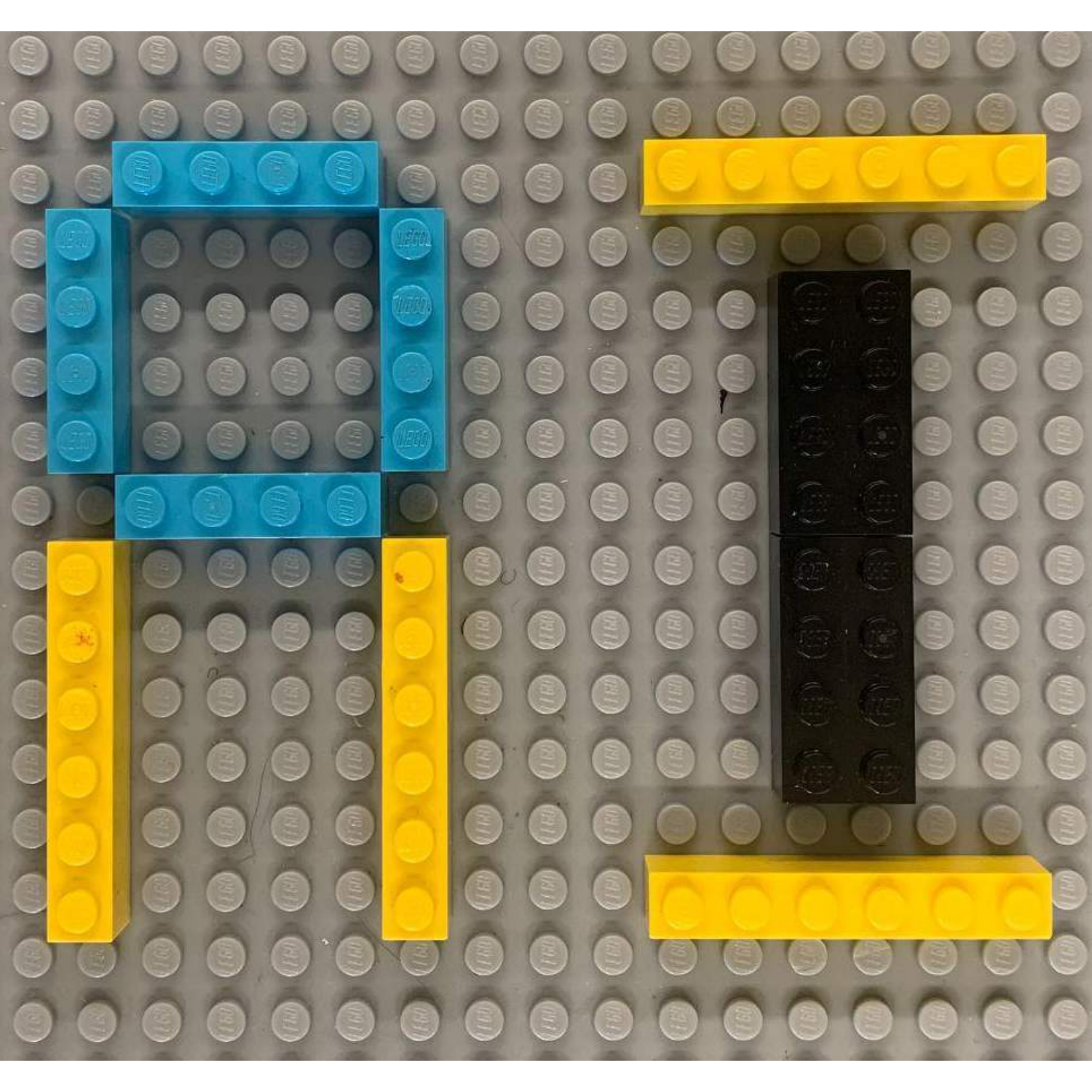}\label{fig:ai}}\hfill
\subfigure[RI.]{\includegraphics[width=0.24\linewidth]{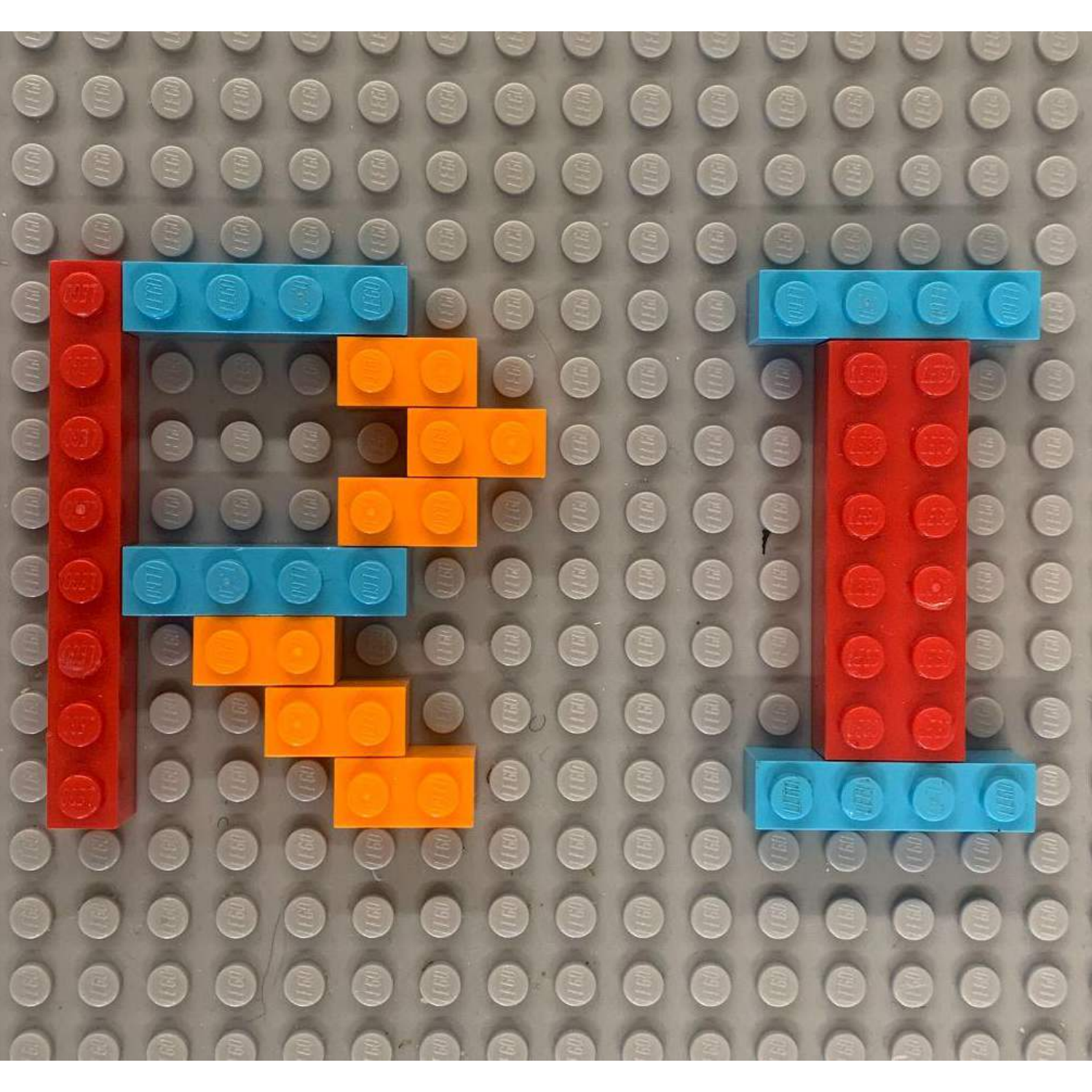}\label{fig:ri}}\hfill
\subfigure[A human.]{\includegraphics[width=0.24\linewidth]{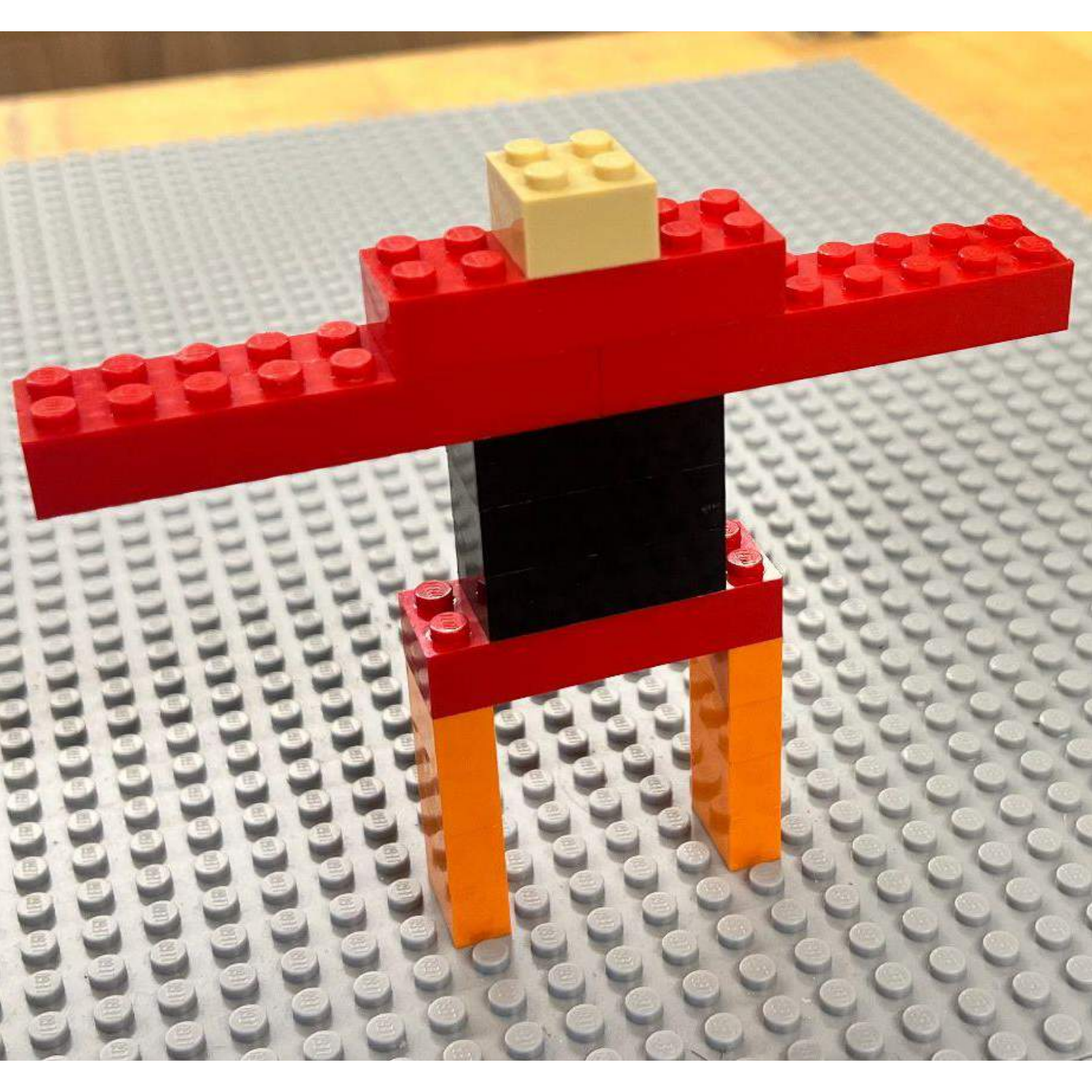}\label{fig:human}}\hfill
\subfigure[A chair.]{\includegraphics[width=0.24\linewidth]{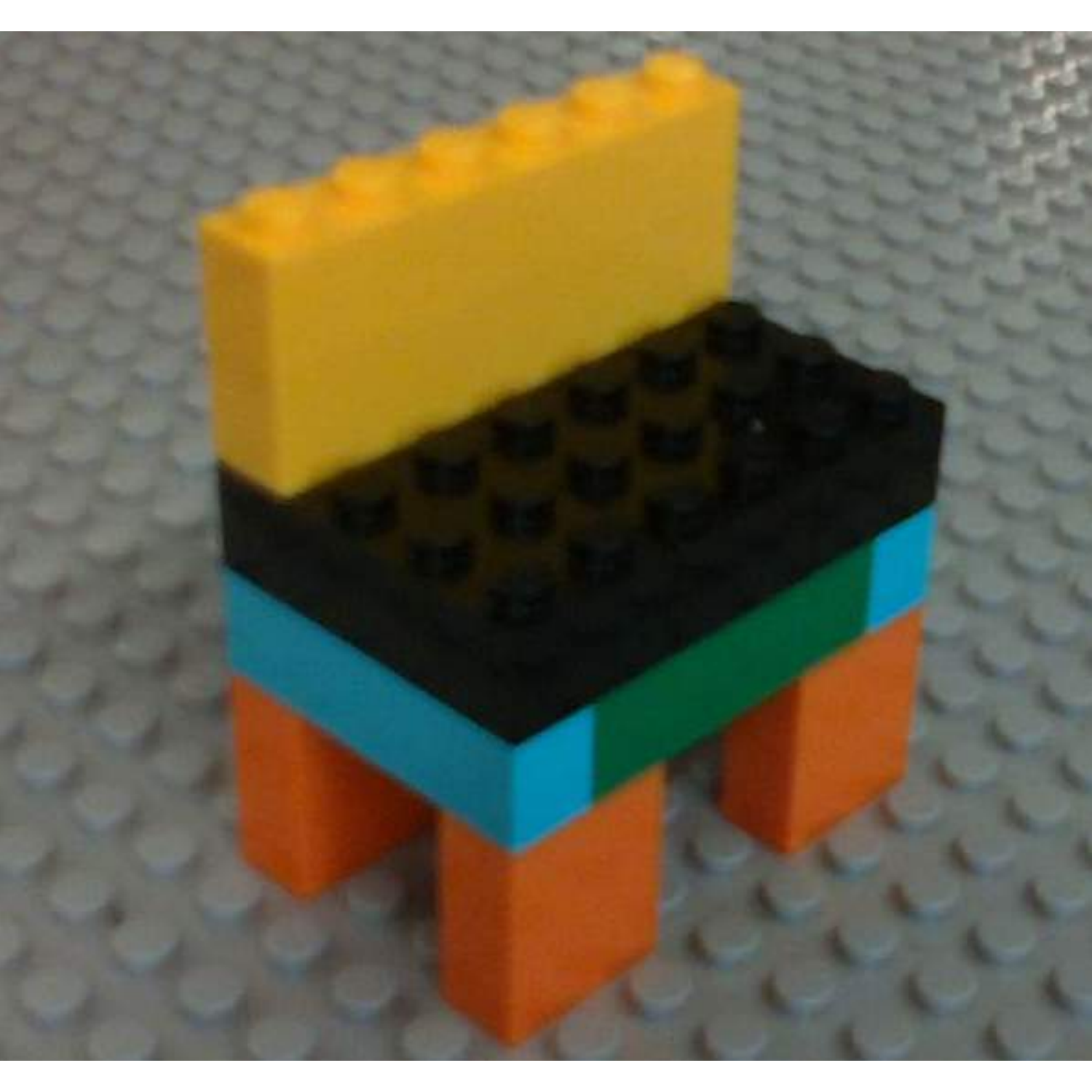}\label{fig:chair}}\\
\vspace{-10pt}
\subfigure[A spiral.]{\includegraphics[width=0.24\linewidth]{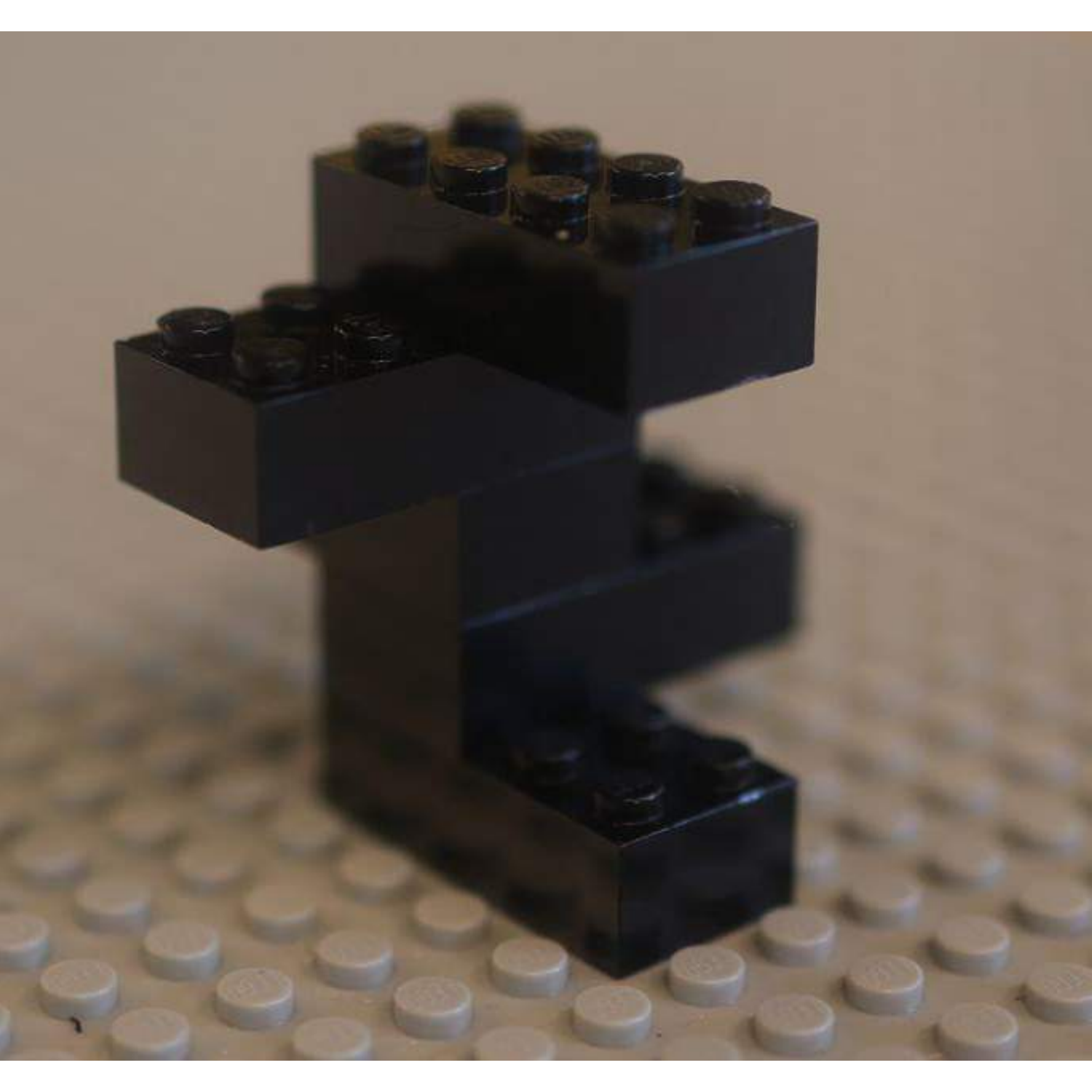}\label{fig:spiral}}\hfill
\subfigure[A bridge.]{\includegraphics[width=0.24\linewidth]{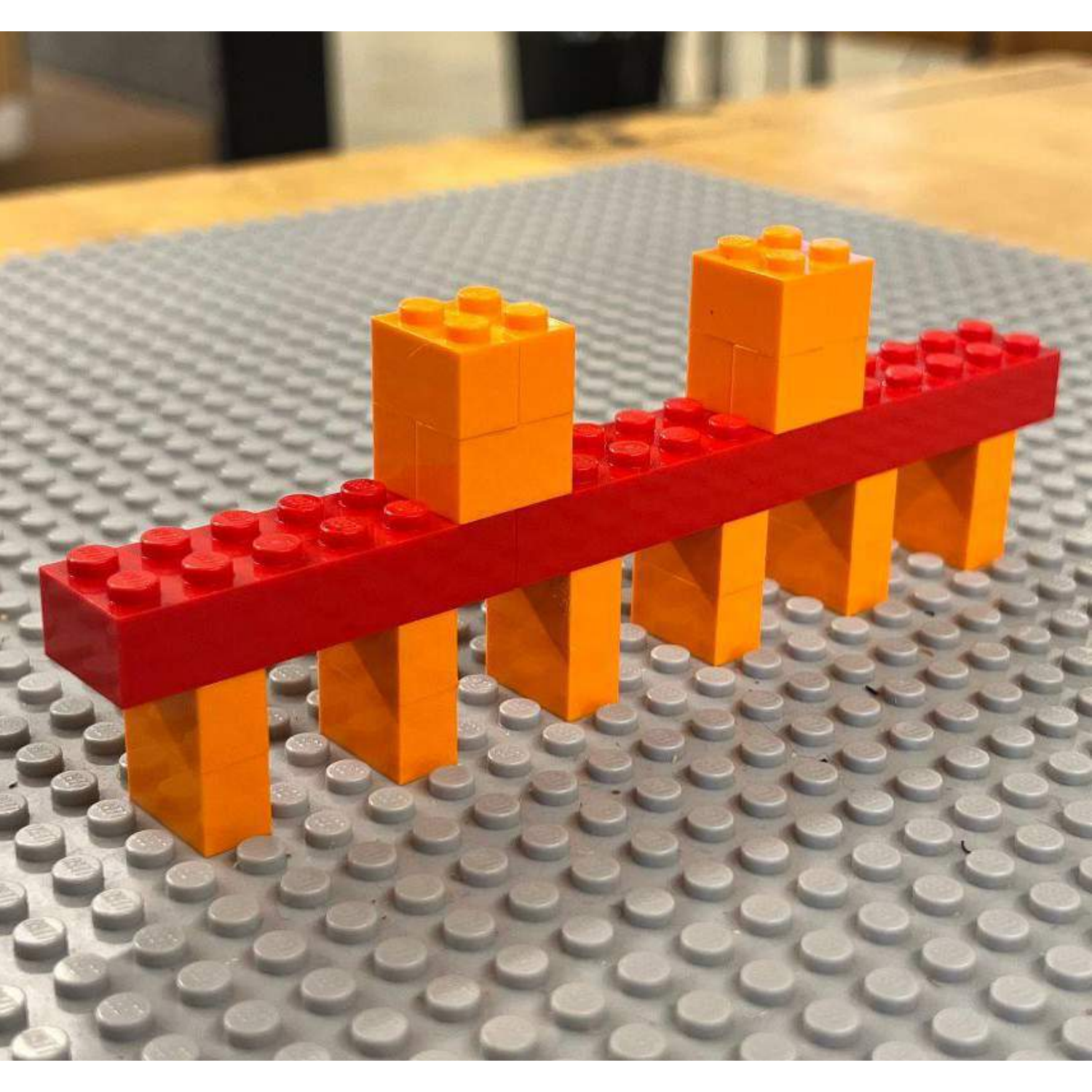}\label{fig:bridge}}\hfill\
\subfigure[A pyramid.]{\includegraphics[width=0.24\linewidth]{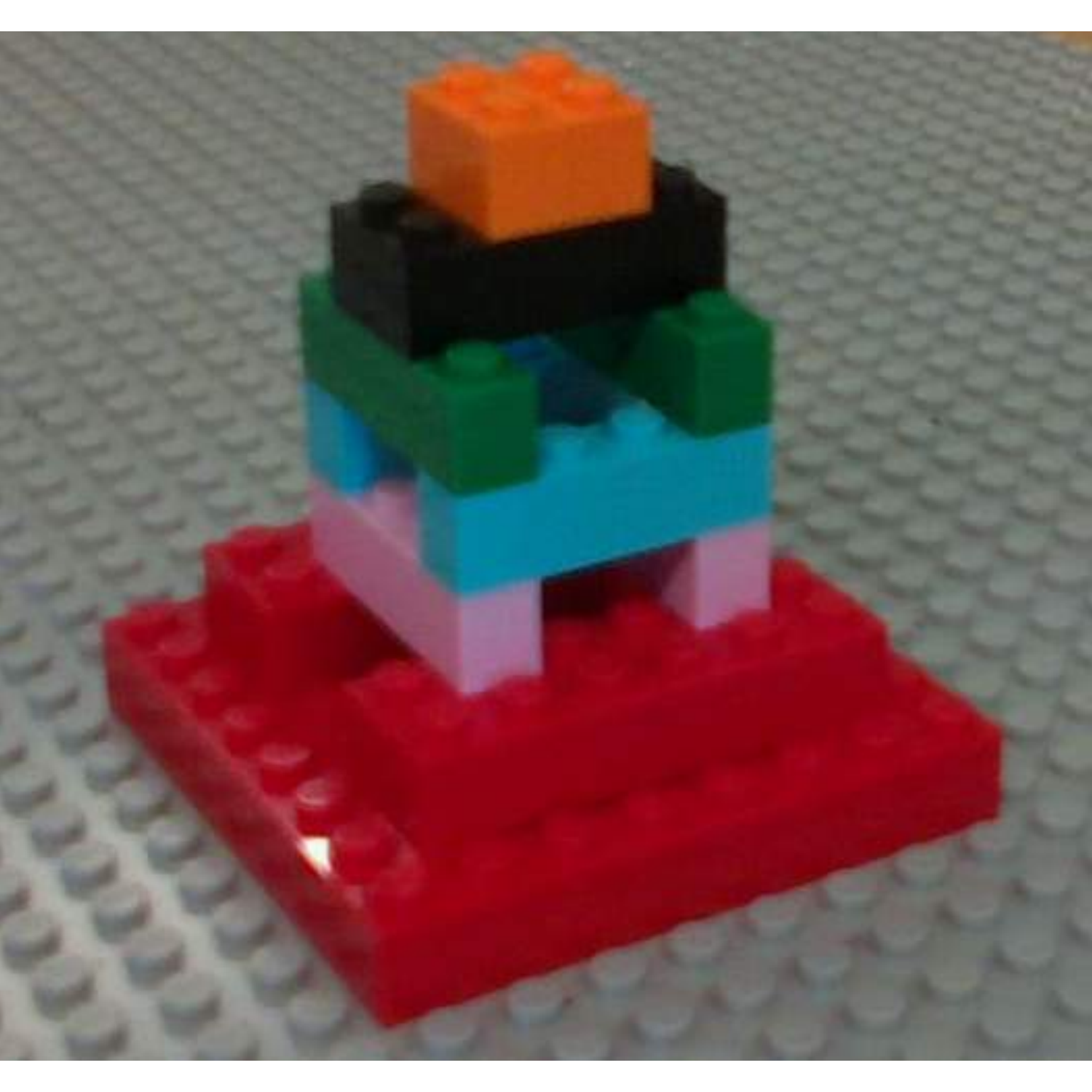}\label{fig:pyramid}}\hfill
\subfigure[A temple.]{\includegraphics[width=0.24\linewidth]{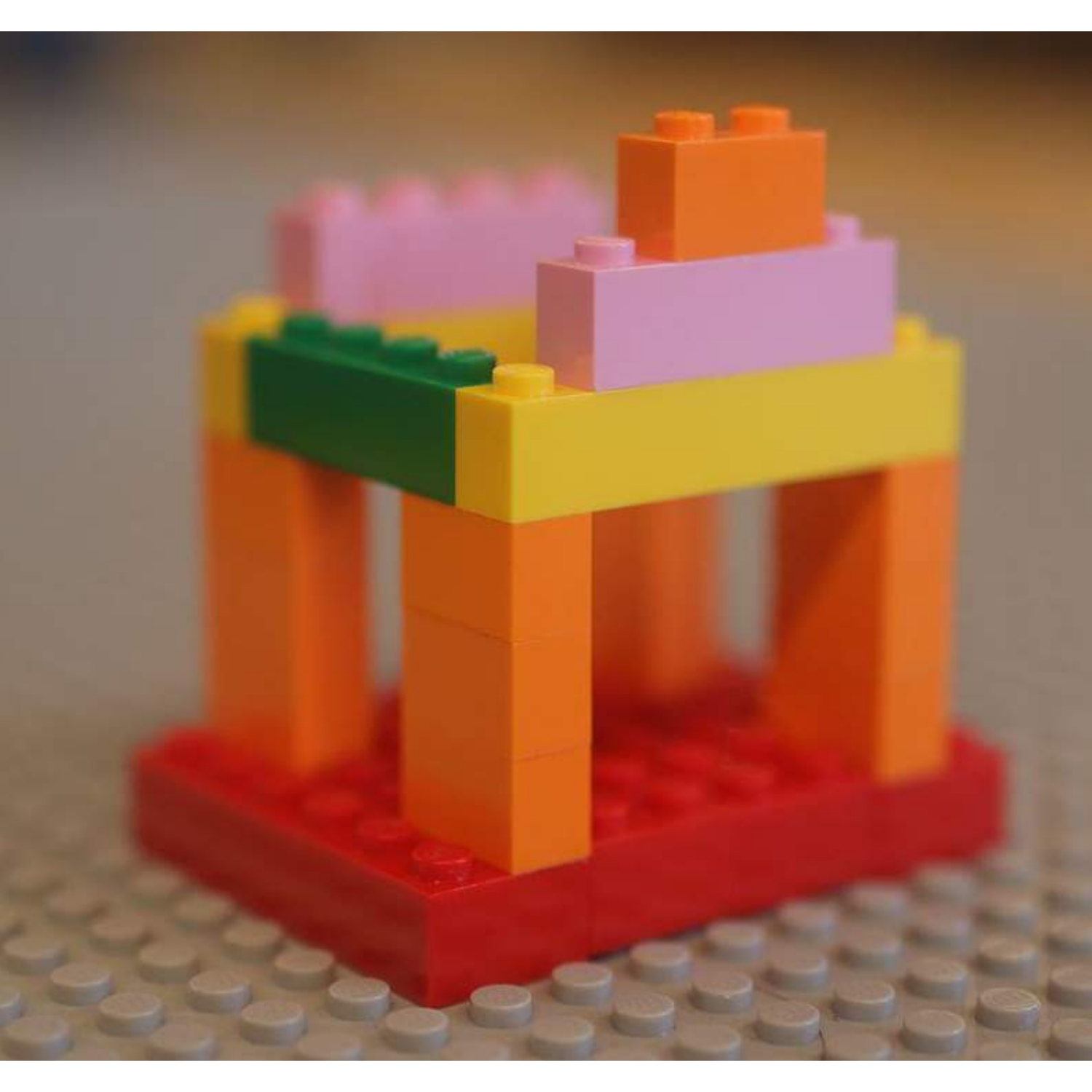}\label{fig:temple}}\\
\vspace{-5pt}
    \caption{\footnotesize Examples of 2D and 3D LEGO prototypes learned from human demonstrations. \label{fig:prototypes}}
    \vspace{-20pt}
\end{figure}

\subsection{Scalability}

\begin{figure*}
\centering
\subfigure[Learning stacking red 1x8 bricks.]{\includegraphics[width=0.33\linewidth]{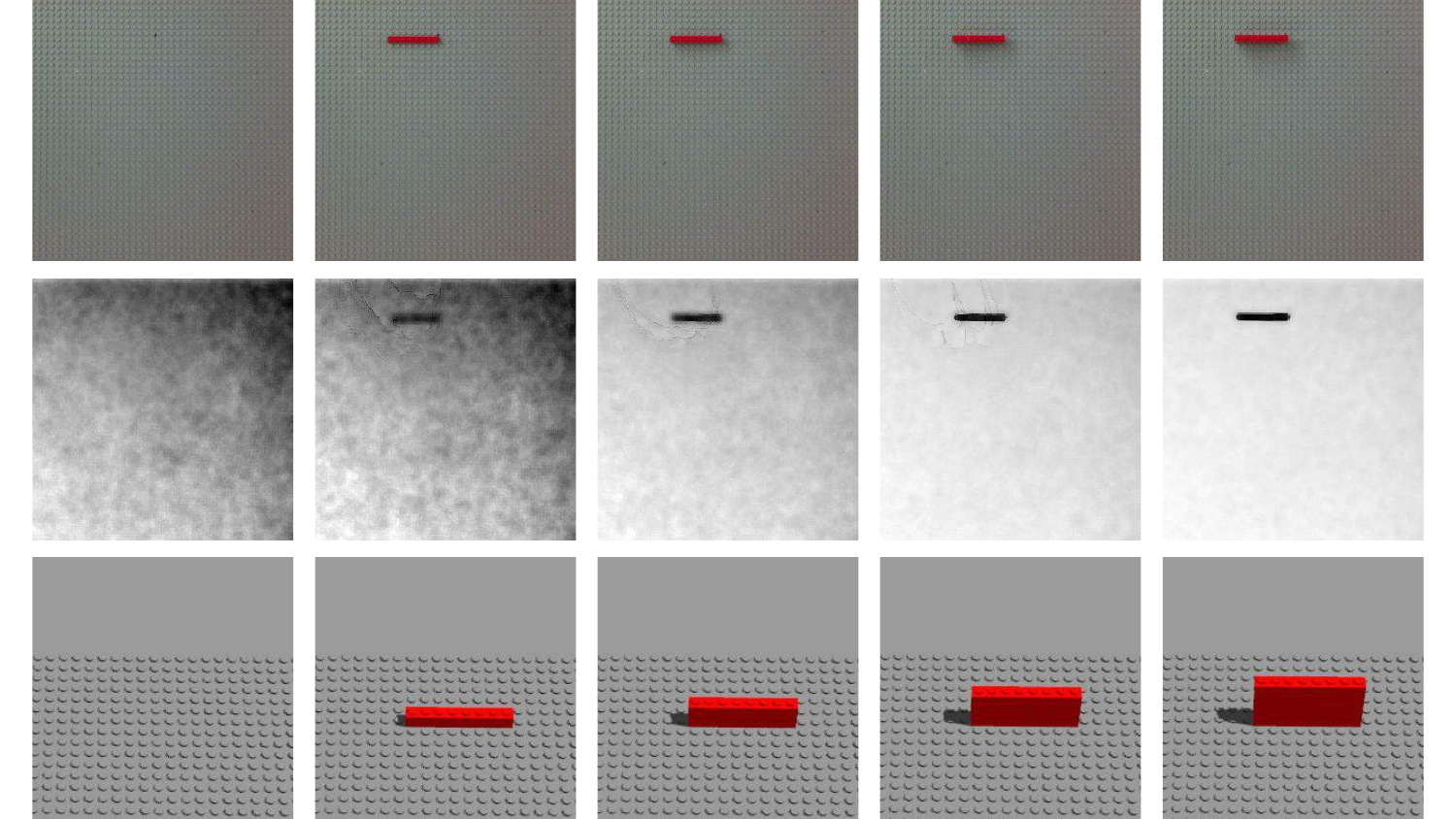}\label{fig:1x8process}}\hfill
\subfigure[Learning stacking red 2x6 bricks.]{\includegraphics[width=0.33\linewidth]{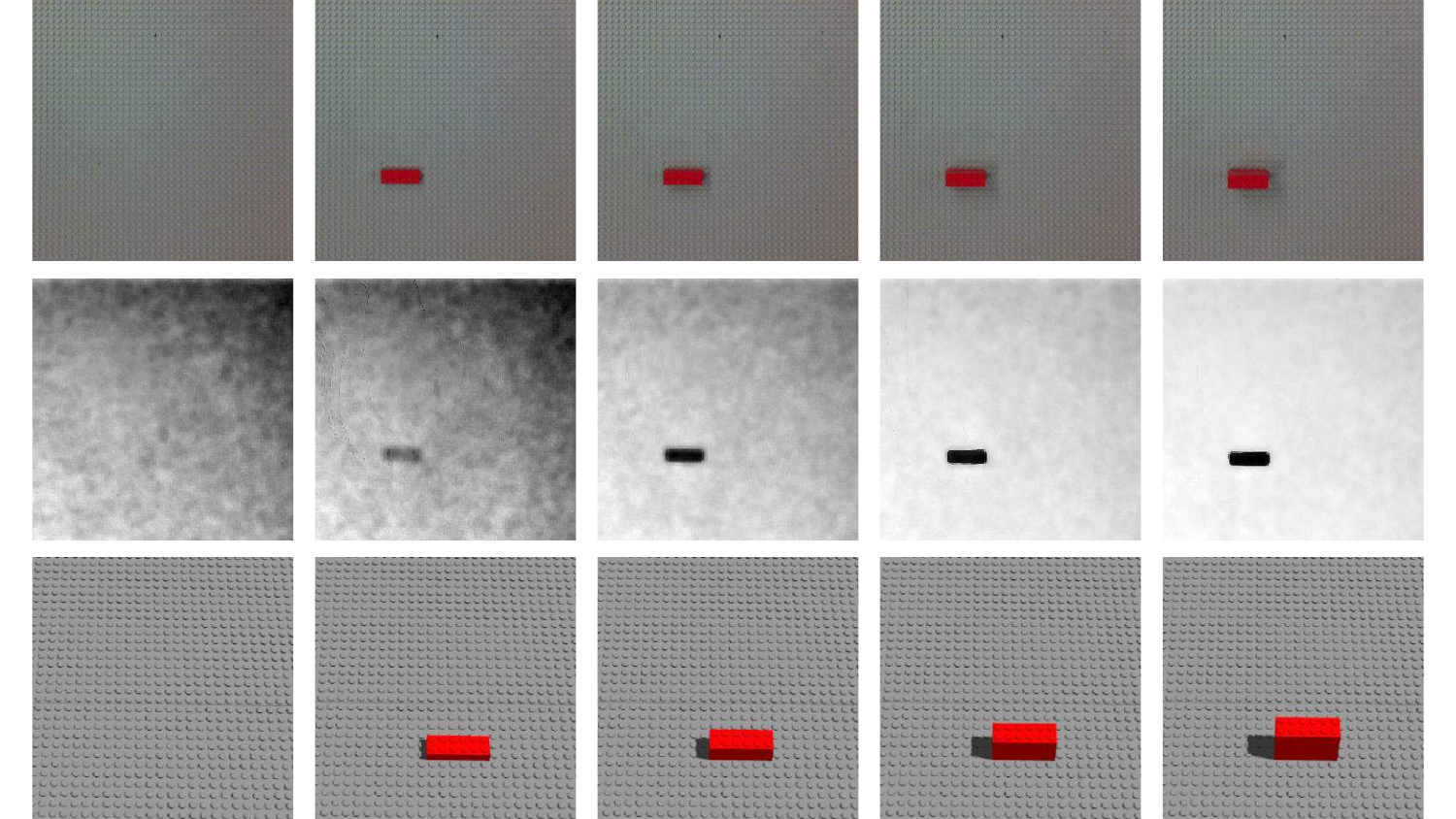}\label{fig:2x6process}}\hfill
\subfigure[Learning stacking yellow 1x6 bricks.]{\includegraphics[width=0.33\linewidth]{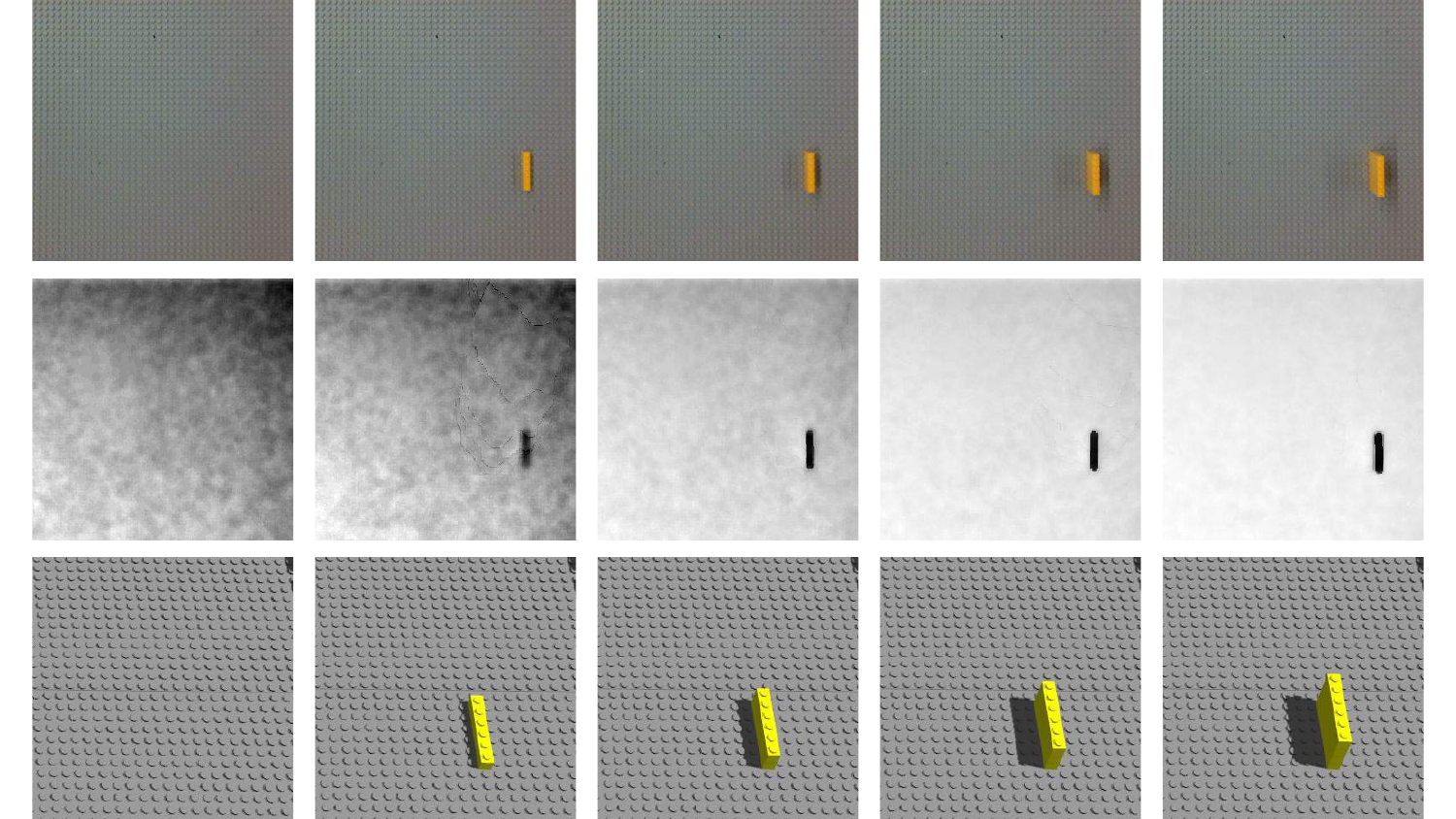}\label{fig:1x6process}}\\
\vspace{-5pt}
    \caption{\footnotesize Learning stacking identical bricks for different bricks. Top: color keyframes of human demonstration. Middle: depth keyframes of human demonstration. Bottom: learned construction visualized in the simulation. \label{fig:learning_stack}}
    \vspace{-20pt}
\end{figure*}

In our experiment, we consider standard LEGO bricks with different dimensions (\ie 1x2, 1x4, 1x6, 1x8, 2x2, 2x4, 2x6) and different colors.
\Cref{fig:learning_stack} illustrates stacking identical bricks straight up. 
We can see that the proposed SaLfD can learn the stacking for different LEGO brick dimensions, and different colors.
In addition, it can also learn the assembly of LEGO structures with different heights (\ie different numbers of layers).
With the integration of the depth channel, the system is robust to color ambiguity and is capable of learning 3D LEGO structures.

\subsection{Robustness}

\begin{figure}
\centering
\subfigure[A spiral.]{\includegraphics[width=0.24\linewidth]{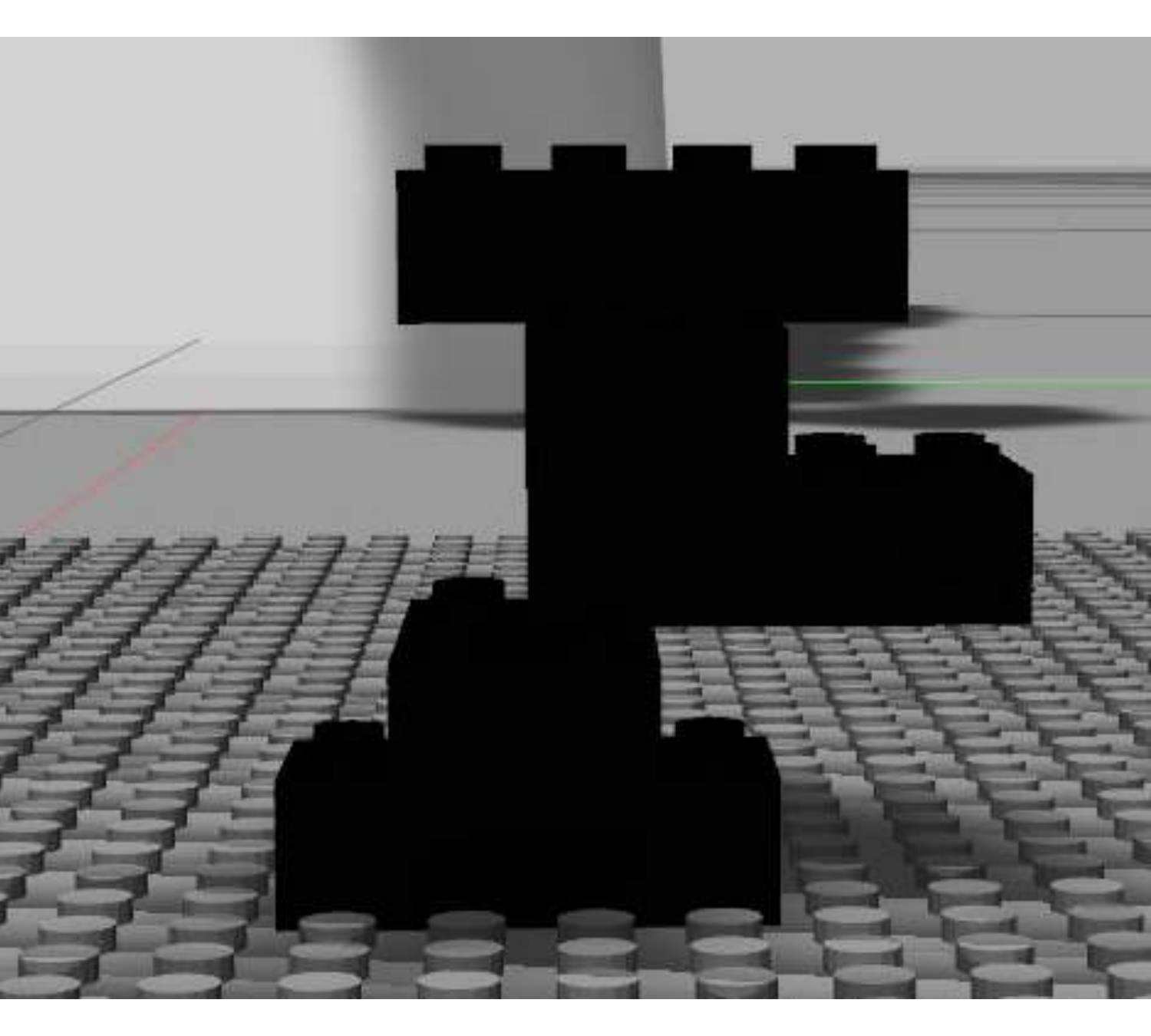}\label{fig:spiral_no_dv}}\hfill
\subfigure[A pyramid.]{\includegraphics[width=0.24\linewidth]{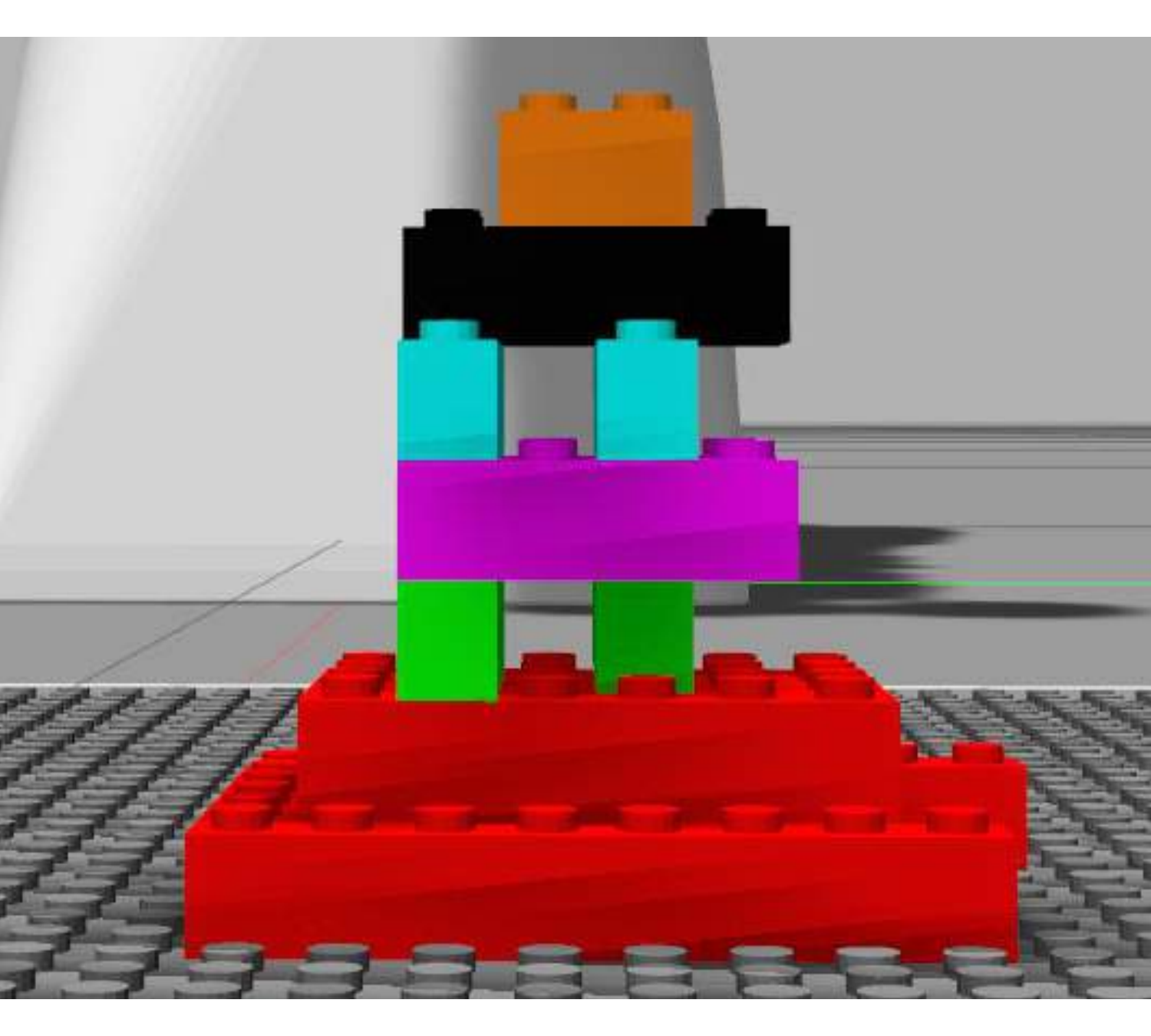}\label{fig:pyramid_no_dv}}\hfill
\subfigure[Stacking blue 1x4.]{\includegraphics[width=0.24\linewidth]{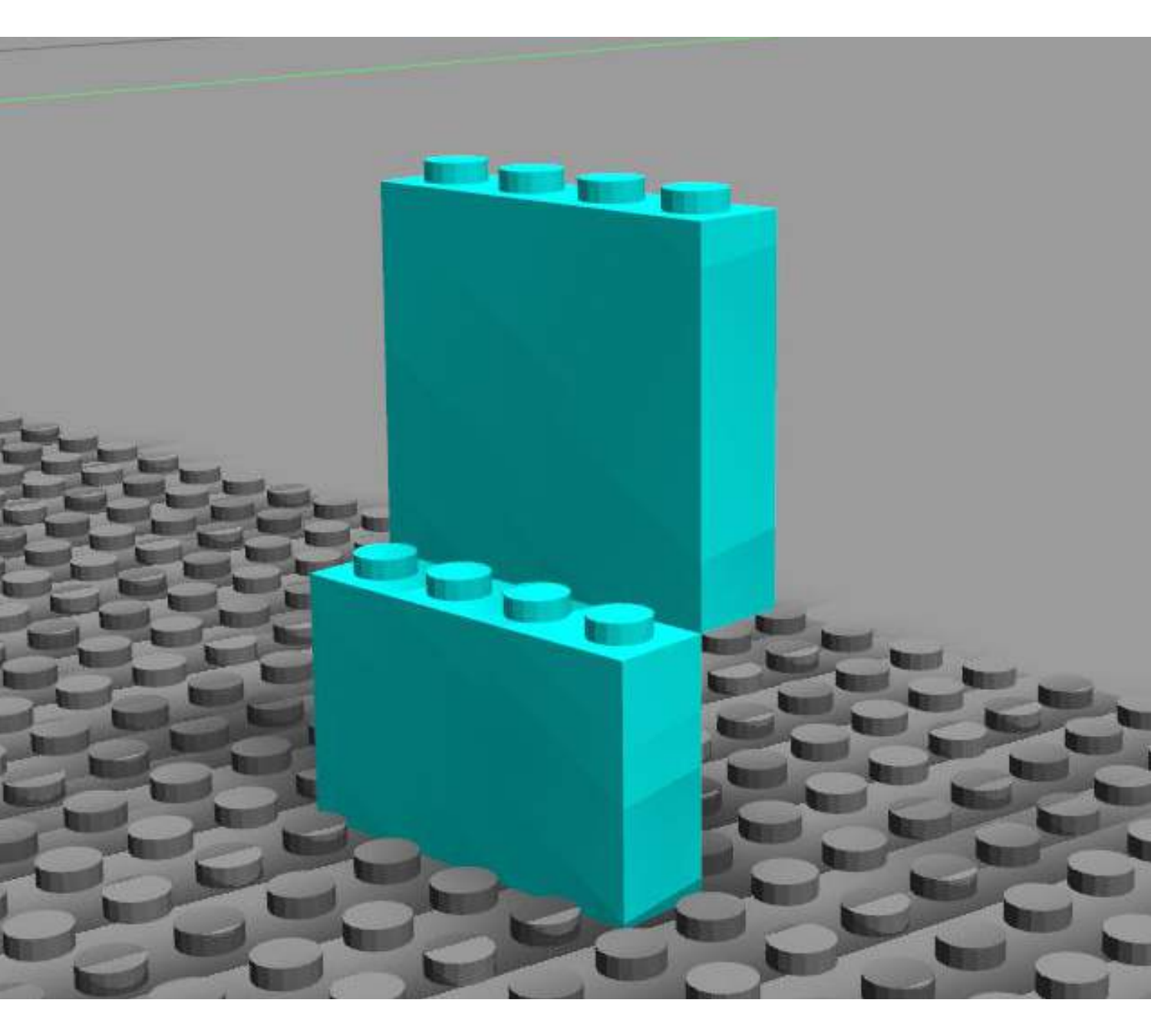}\label{fig:stack_blue_no_dv}}\hfill
\subfigure[Stacking pink 1x4.]{\includegraphics[width=0.24\linewidth]{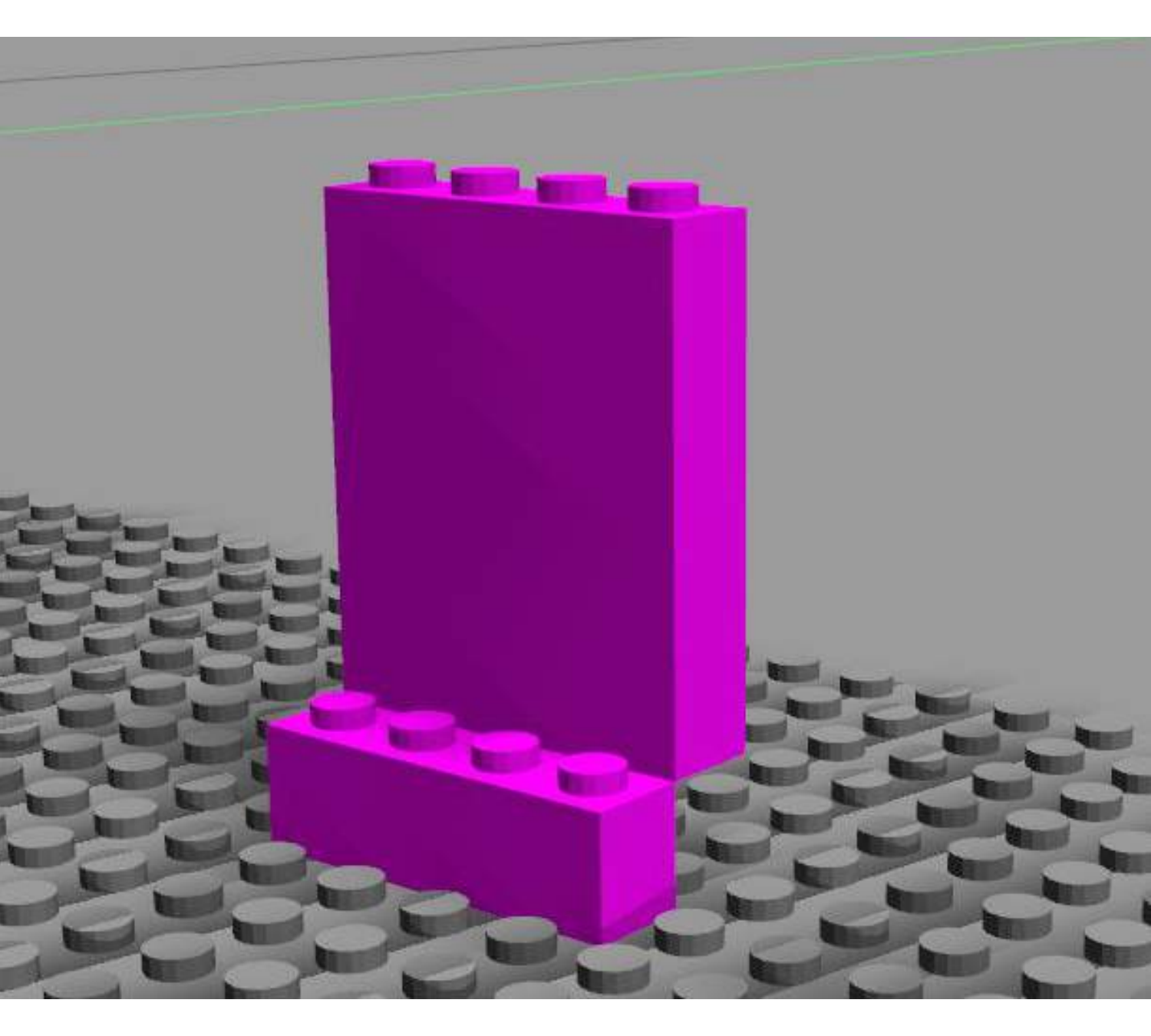}\label{fig:stack_pink_no_dv}}
\\
\vspace{-10pt}
\subfigure[A spiral.]{\includegraphics[width=0.24\linewidth]{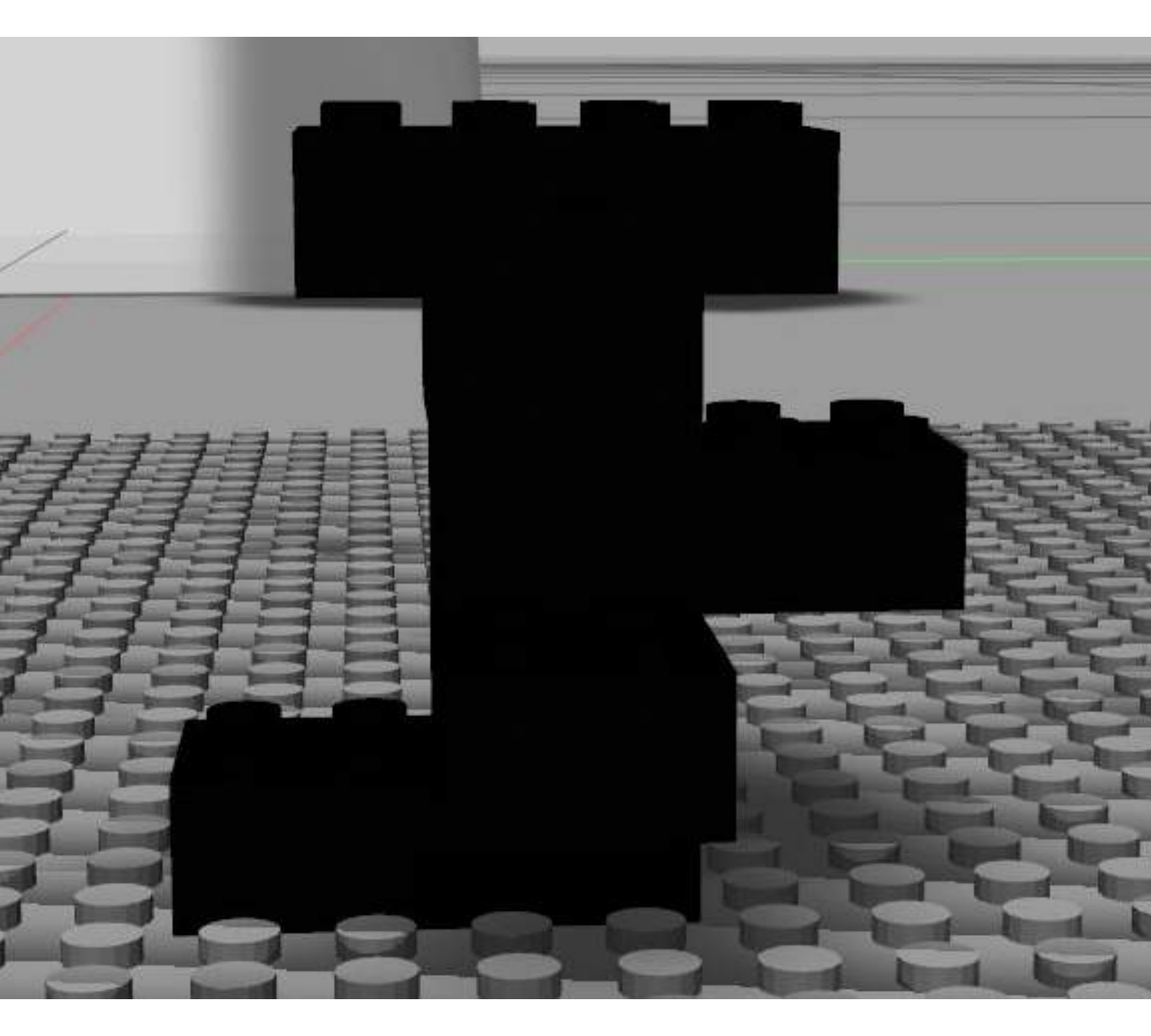}\label{fig:spiral_dv}}\hfill
\subfigure[A pyramid.]{\includegraphics[width=0.24\linewidth]{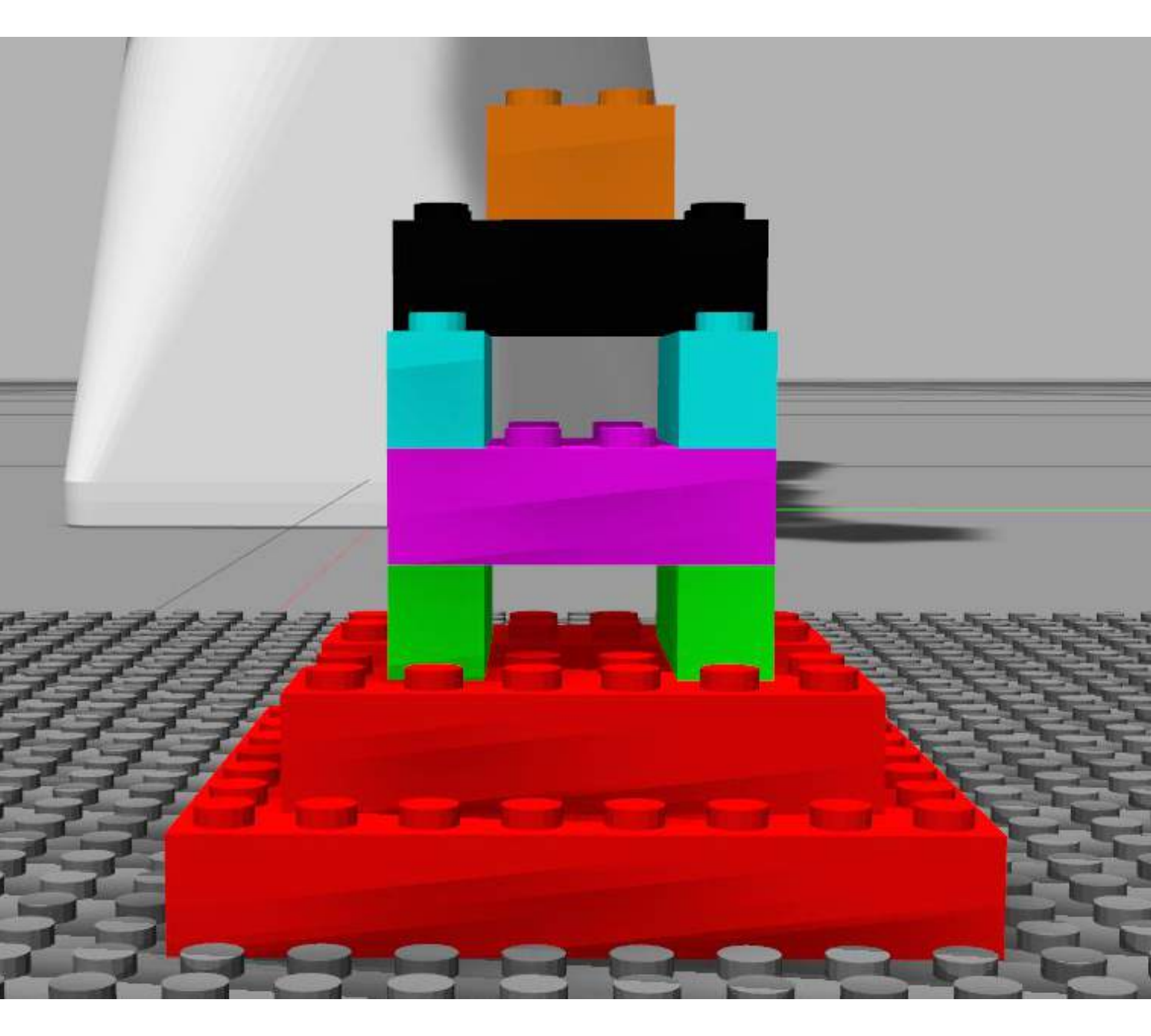}\label{fig:pyramid_dv}}\hfill
\subfigure[Stacking blue 1x4.]{\includegraphics[width=0.24\linewidth]{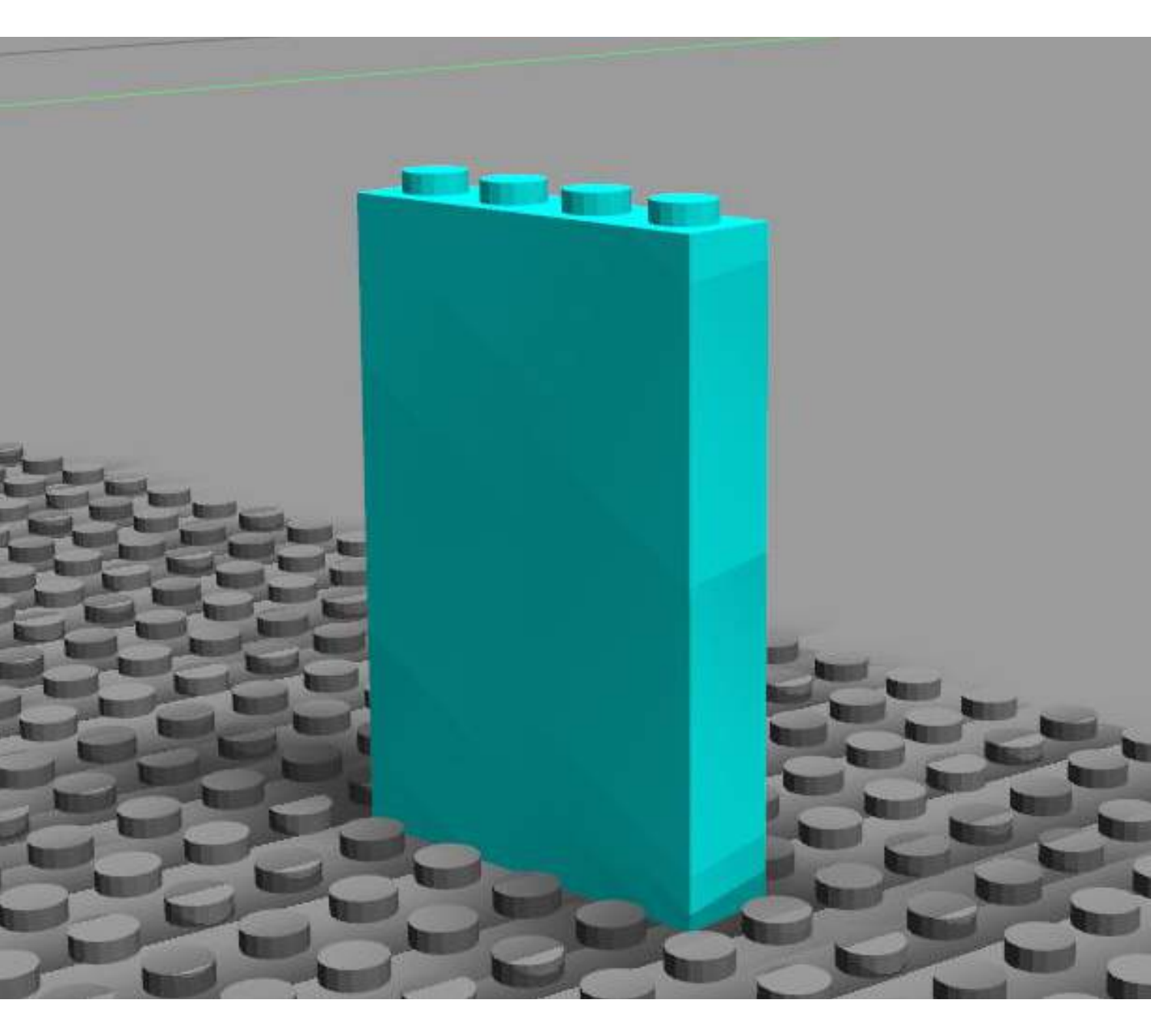}\label{fig:stack_blue_dv}}\hfill
\subfigure[Stacking pink 1x4.]{\includegraphics[width=0.24\linewidth]{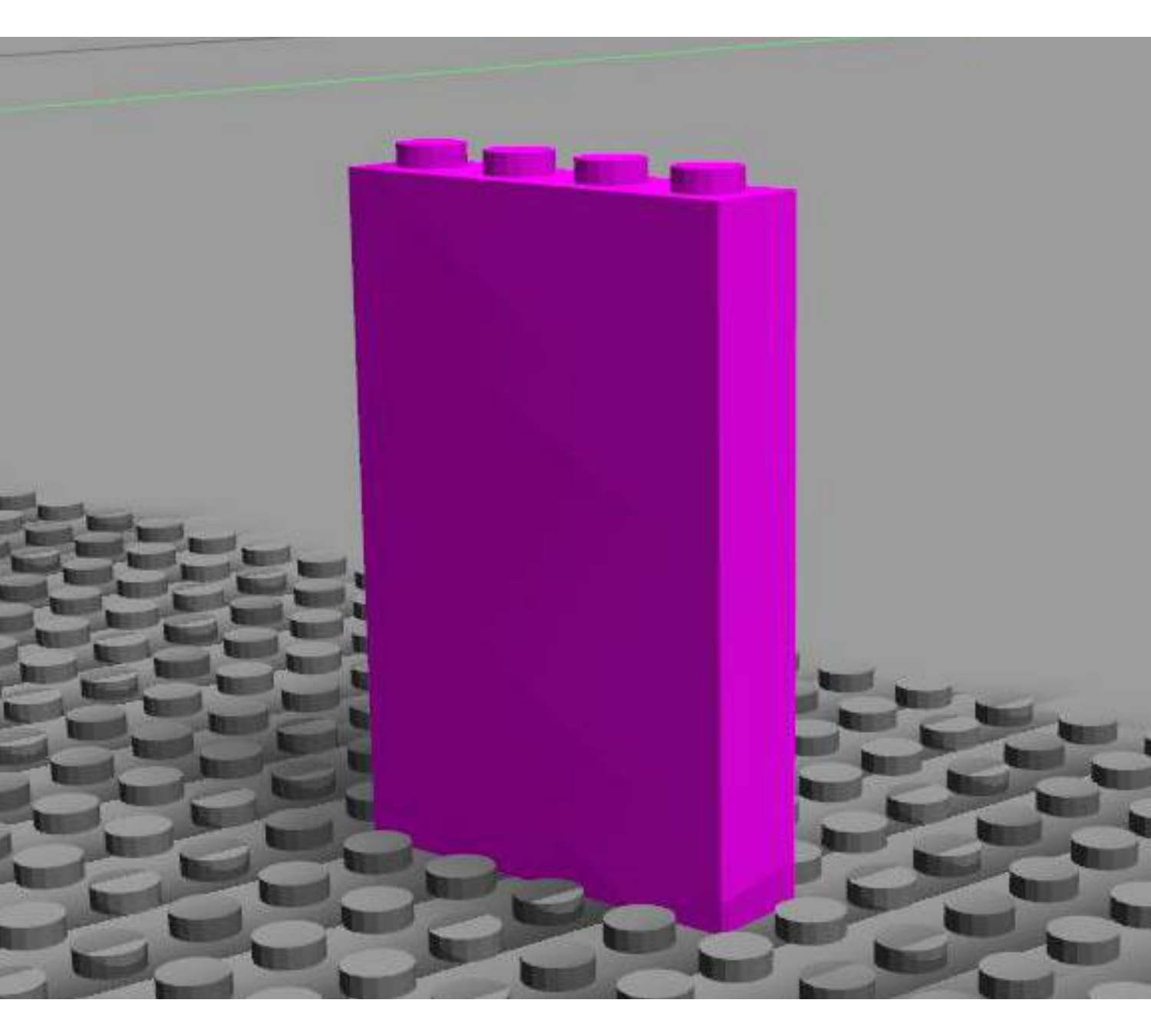}\label{fig:stack_pink_no_dv}}
\vspace{-5pt}
    \caption{\footnotesize Comparisons of the learned LEGO constructions. Top: without verification. Bottom: with simulation verification. \label{fig:exp2}}
    \vspace{-20pt}
\end{figure}

To demonstrate the advantage of the simulation verification in SaLfD, we compare the learned LEGO structures with and without the simulation verification.
\Cref{table:comparison_dv} quantitatively compares the learning success rate.
We also include the rule-based method in \cite{liu2023robotic} with a single visual camera input as the baseline.
Each object is built for 10 trials.
We can see the baseline performs decently on the 2D structures, but it cannot handle 3D LEGO structures.
Also, we can see the proposed SaLfD is more robust in terms of the success rate due to the simulation verification.
\Cref{fig:exp2} shows examples of the learned LEGO structures with and without the verification.
Without the verification, the generated LEGO structure is close to the human-intended structure (despite the low success rate), but with slight deviation due to the measurement uncertainty in \cref{fig:Challenges}.
This indicates that the neural network models are able to extract the features, but are not able to generalize well to exactly match the human needs as shown in \cref{fig:constraints}.
On the other hand, the simulation verification is able to correct the errors and generate a LEGO construction plan that represents the identical target object.
This indicates that with the assistance of the simulation verification, the pipeline is more robust against the uncertainty.


\begin{table}
\centering
\begin{tabular}{c  c c  c   } 
\hline
LEGO Prompt & Baseline \cite{liu2023robotic} & LfD & SaLfD  \\ 
\hline
Characters: AI &  90\%  &  73\% & \textbf{100\%}\\
Characters: RI &  87.5\% & 70.8\% &  \textbf{96.7\%} \\
Human  &  NA  & 67.6\% &  \textbf{87.1\%} \\
Chair  & NA  &  69\%  & \textbf{94.3\%} \\
Spiral   &  NA  &  71\% & \textbf{98\%} \\
Bridge &   NA   &  63.15\% & \textbf{86.3\%} \\
Pyramid &  NA &   76.7\% & \textbf{100\%} \\
Temple &  NA  &  63.5\%  &    \textbf{95.2\%} \\
\hline
\end{tabular}
\caption{\footnotesize Comparison of the success rate of the learned LEGO structures of different methods. LfD is the one without simulation verification. \label{table:comparison_dv}}
\vspace{-20pt}
\end{table}

\subsection{Efficiency}

\begin{table}
\centering
\begin{tabular}{c  c  c  c   } 
\hline
LEGO Prompt & Handcrafted & Simulation & SaLfD  \\    
\hline
Characters: AI \{10\} &  168.6 (16.8)  &  302.3 (30.2) & \textbf{66.2 (6.6)}\\
Characters: RI \{12\} &  213.0 (17.7) &  369.4 (30.7) &\textbf{72.6 (6.0)}\\
Human \{17\} &  249.2 (14.6) &  529.4 (31.1) & \textbf{101.6 (5.9)}  \\
Chair \{21\} & 305.9 (14.5) &  694.7 (33.0) & \textbf{127.5 (6.0)}\\
Spiral \{5\}  &  64.7 (12.9) &  145.9 (29.1) &\textbf{23.6 (4.7)} \\
Bridge \{19\} &   262.9 (13.8)  & 640.4 (33.7) &\textbf{126.4 (6.6)}\\
Pyramid \{15\} &  264.0 (17.6) & 464.2 (30.9)  &\textbf{91.6 (6.1)} \\
Temple \{23\} &   333.4 (14.5) & 790.5 (34.3) &\textbf{133.5 (5.8)} \\
\hline
\end{tabular}
\caption{\footnotesize Comparison of the average time (\si{\second}) taken for generating a LEGO construction plan. $\{\cdot\}$ indicates the number of bricks used to build the prototype. $(\cdot)$ shows the time per brick. \label{table:comparison_eff}}
\vspace{-20pt}
\end{table}

\begin{figure*}
\centering
\subfigure[]{\includegraphics[width=0.16\linewidth]{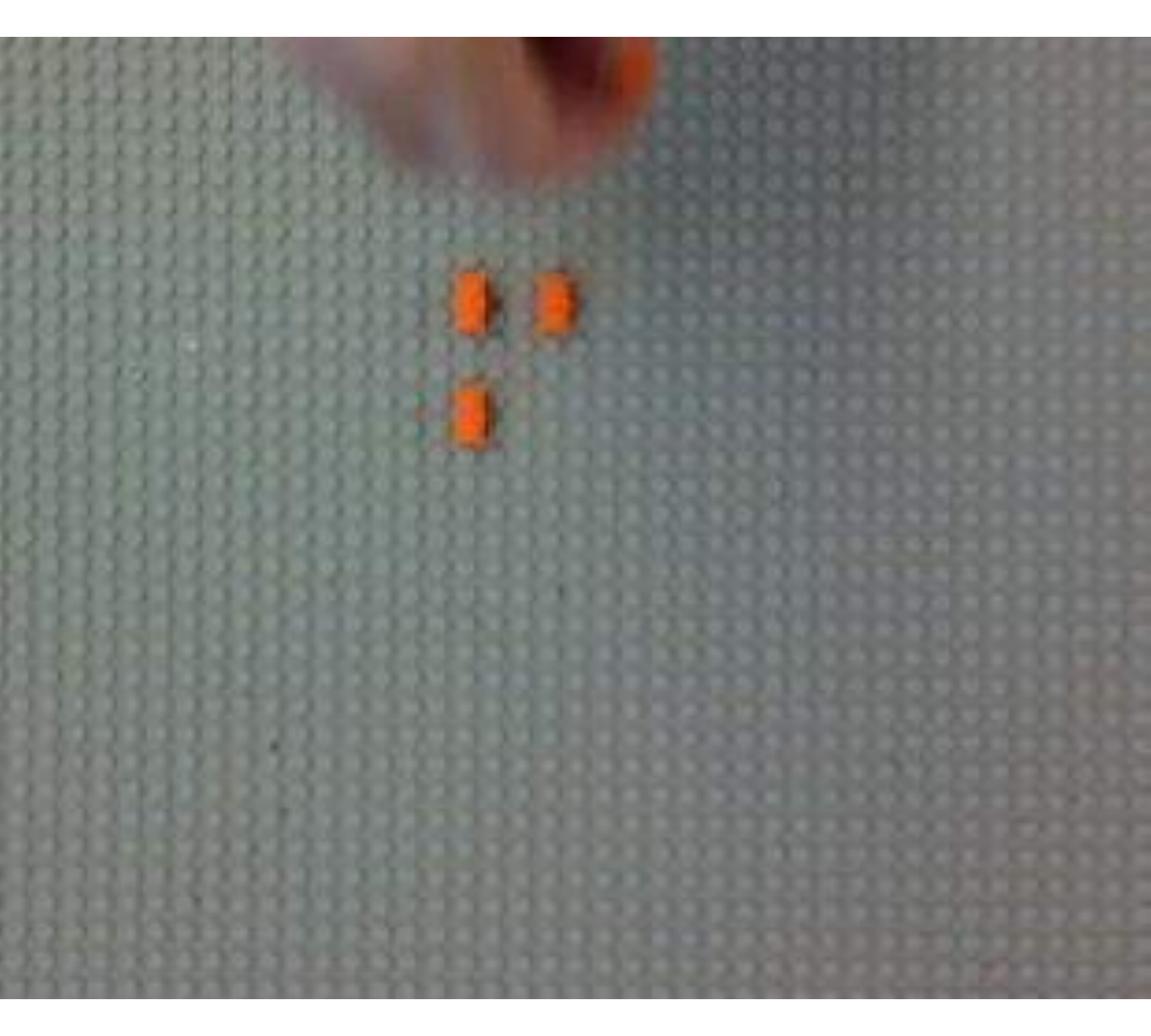}\label{fig:h1}}\hfill
\subfigure[]{\includegraphics[width=0.16\linewidth]{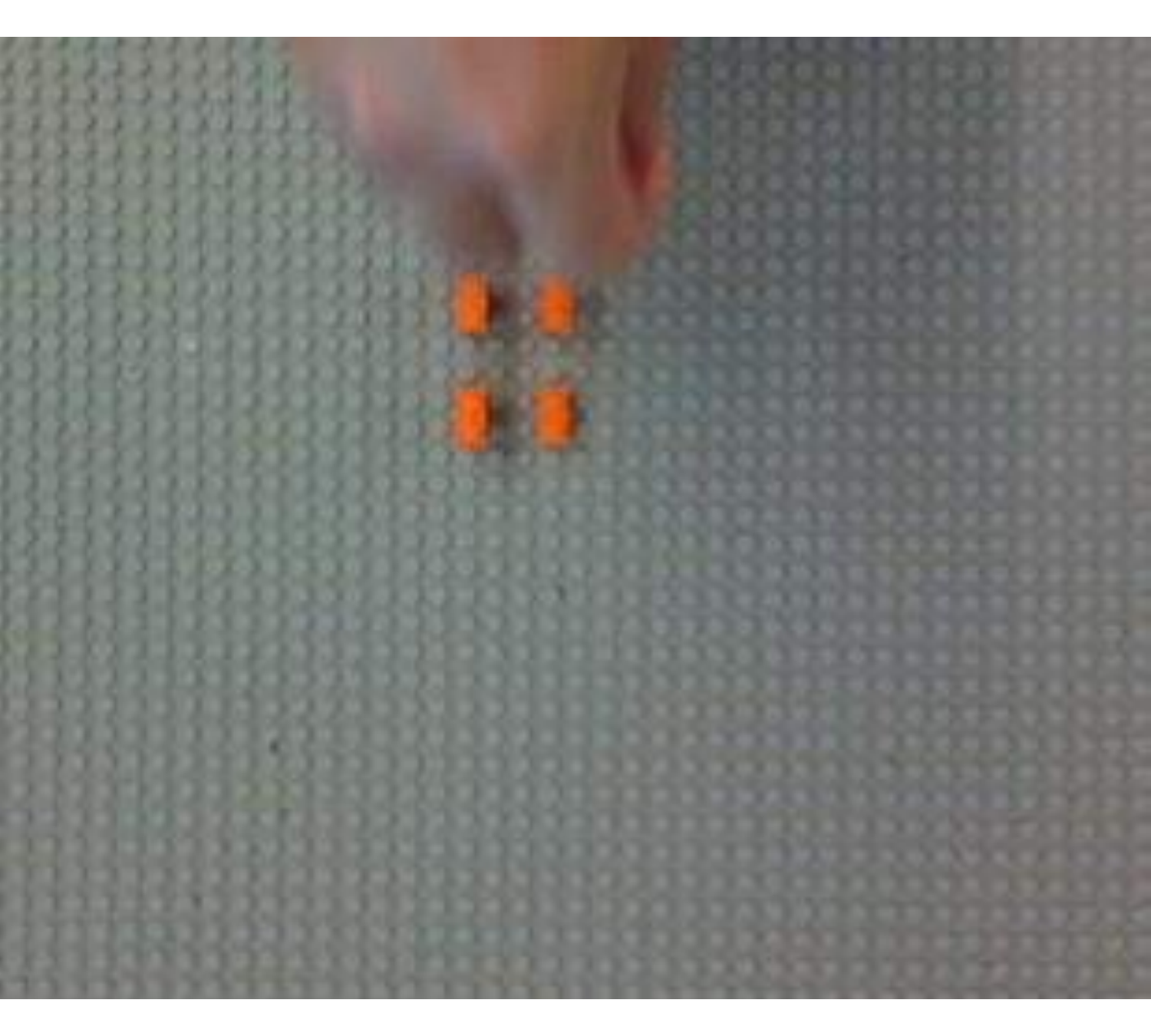}\label{fig:h2}}\hfill
\subfigure[]{\includegraphics[width=0.16\linewidth]{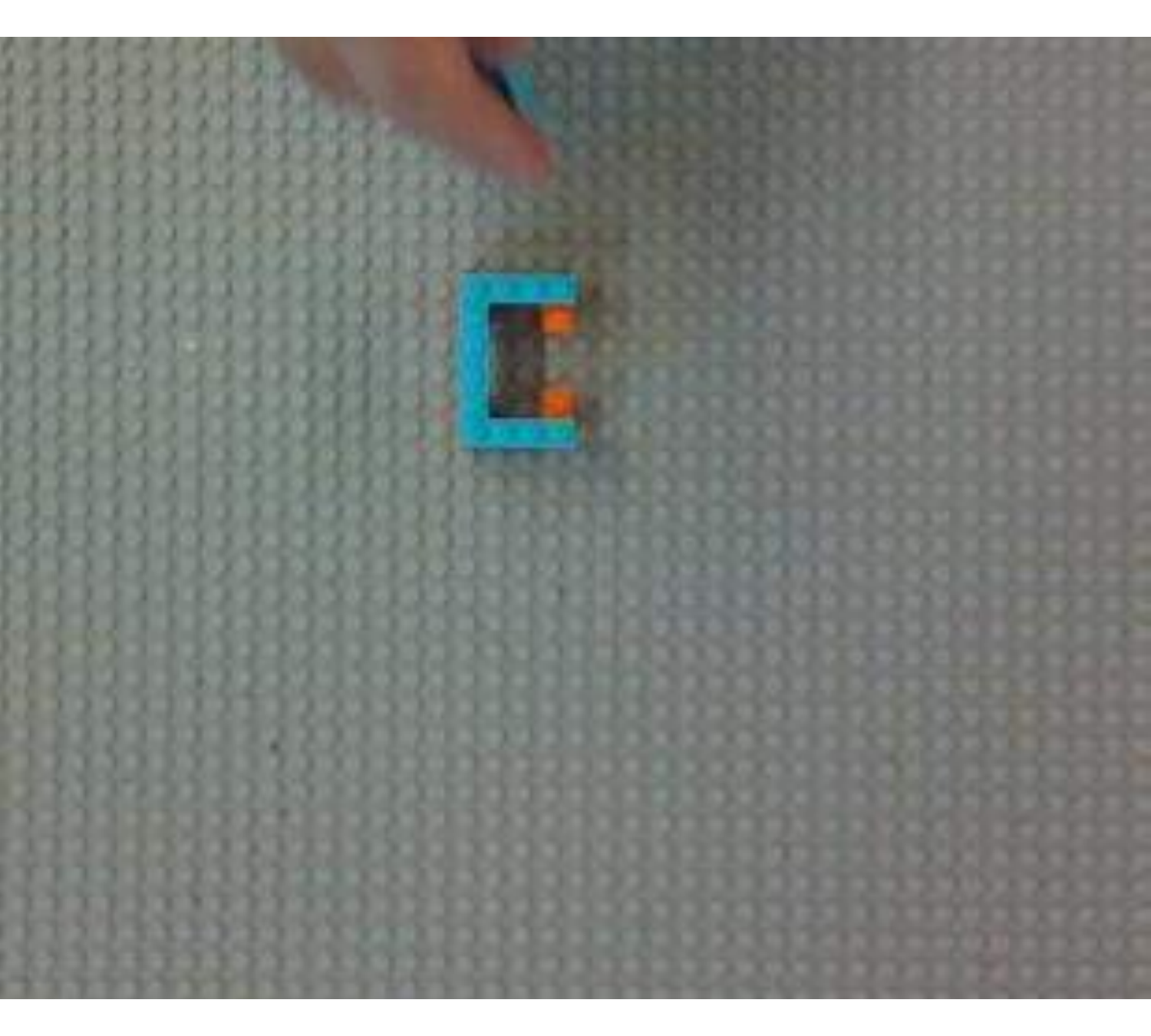}\label{fig:h3}}\hfill
\subfigure[]{\includegraphics[width=0.16\linewidth]{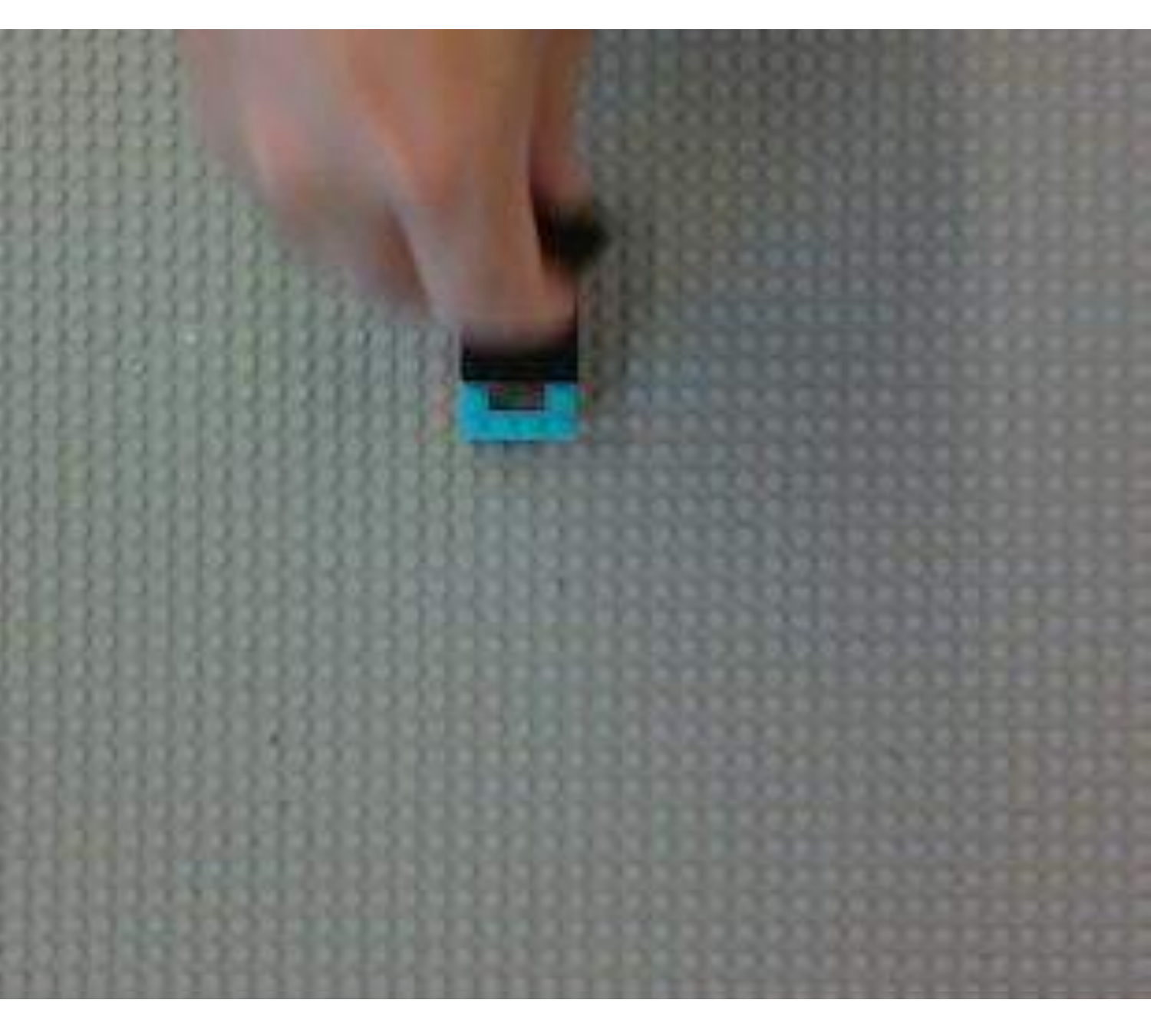}\label{fig:h4}}\hfill
\subfigure[]{\includegraphics[width=0.16\linewidth]{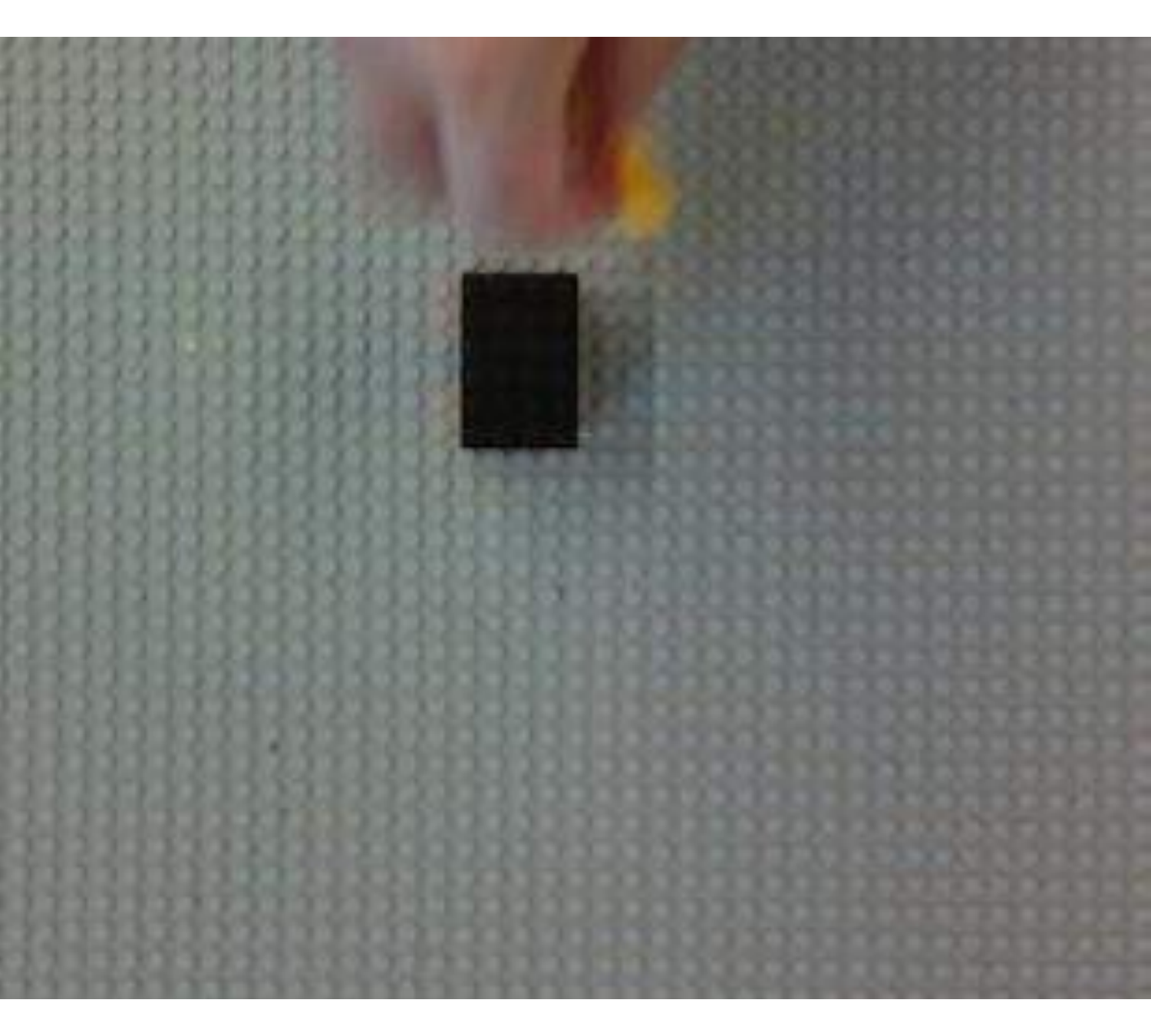}\label{fig:h5}}\hfill
\subfigure[]{\includegraphics[width=0.16\linewidth]{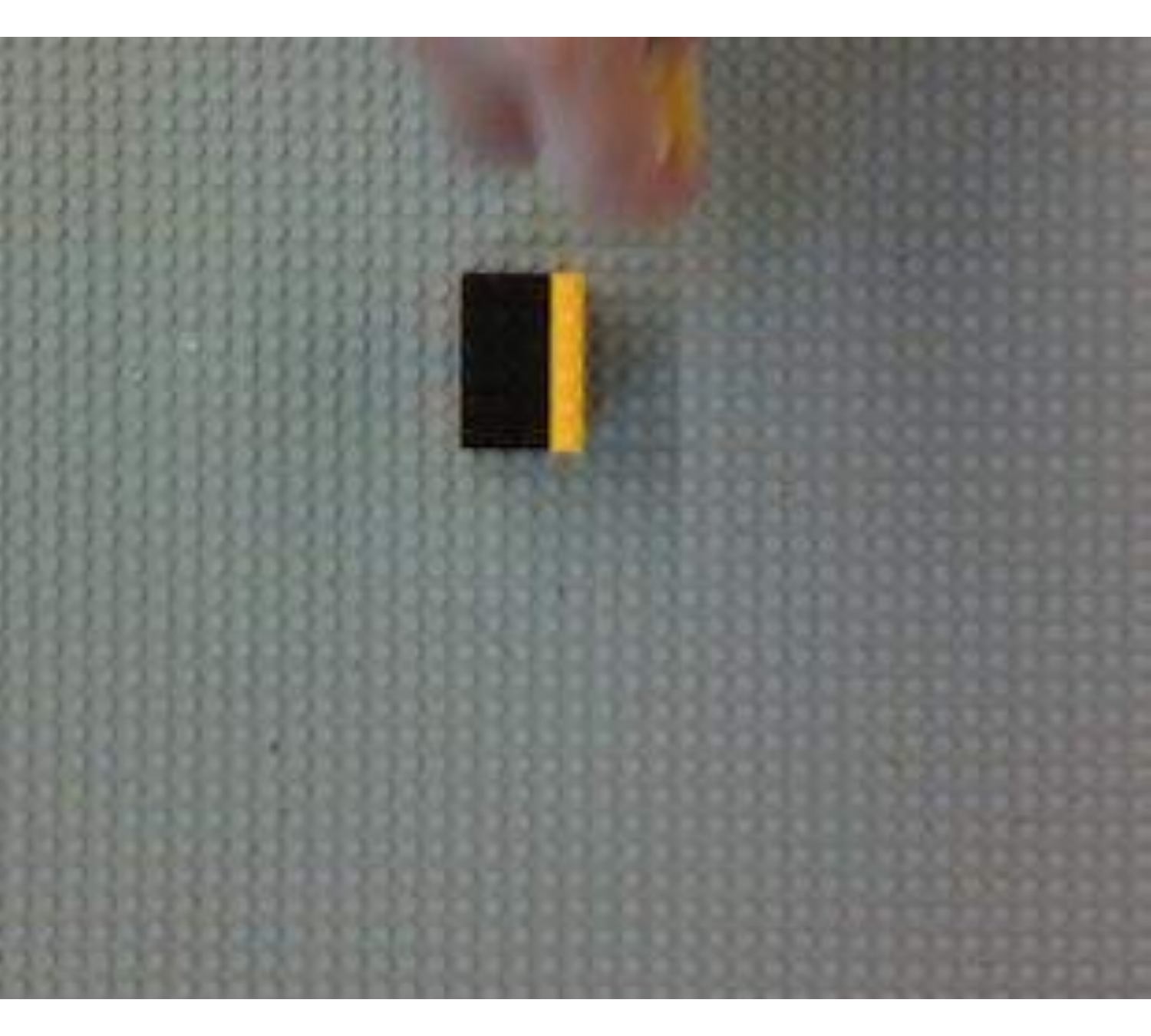}\label{fig:h6}}
\\
\vspace{-10pt}
\subfigure[]{\includegraphics[width=0.16\linewidth]{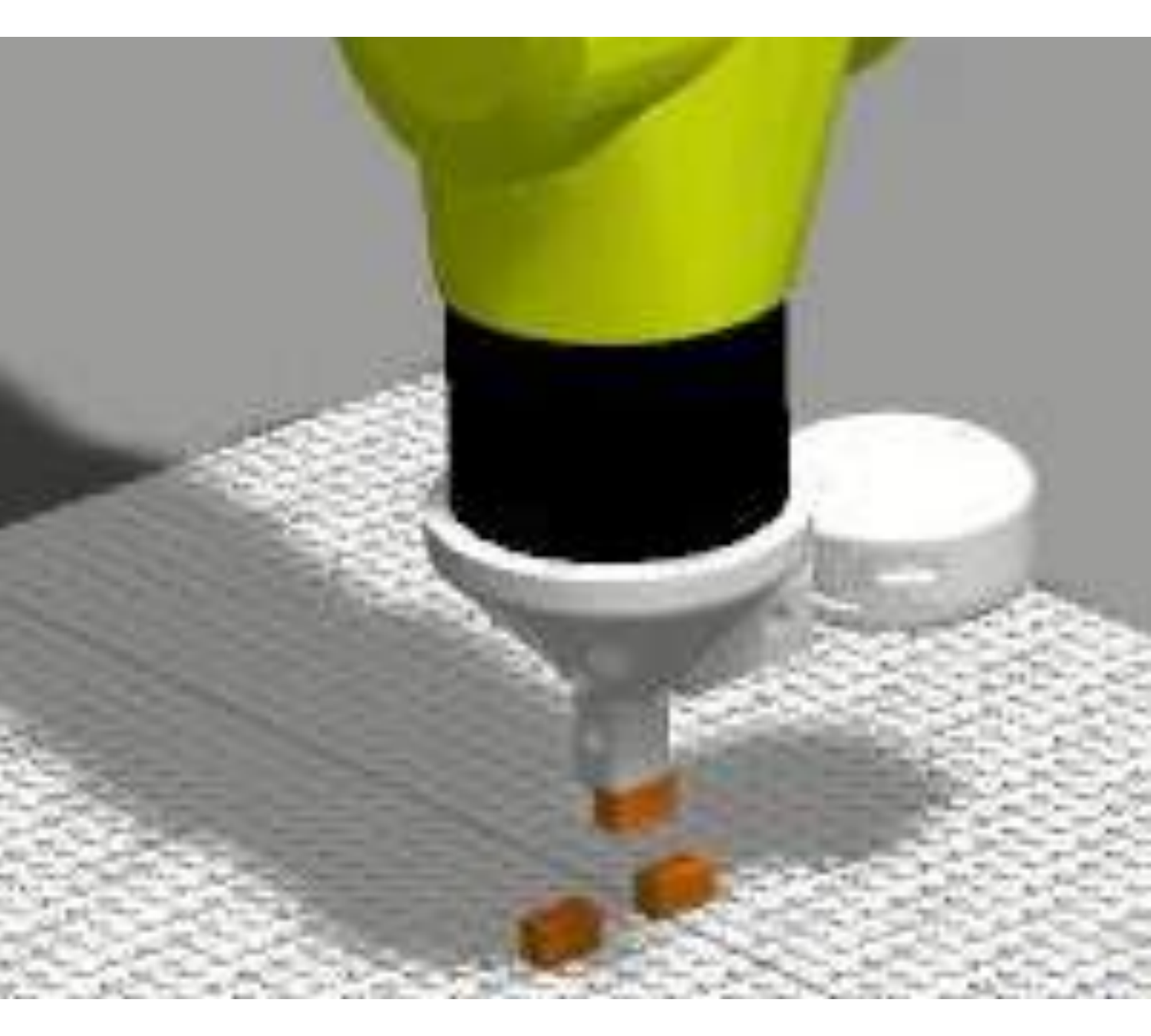}\label{fig:1}}\hfill
\subfigure[]{\includegraphics[width=0.16\linewidth]{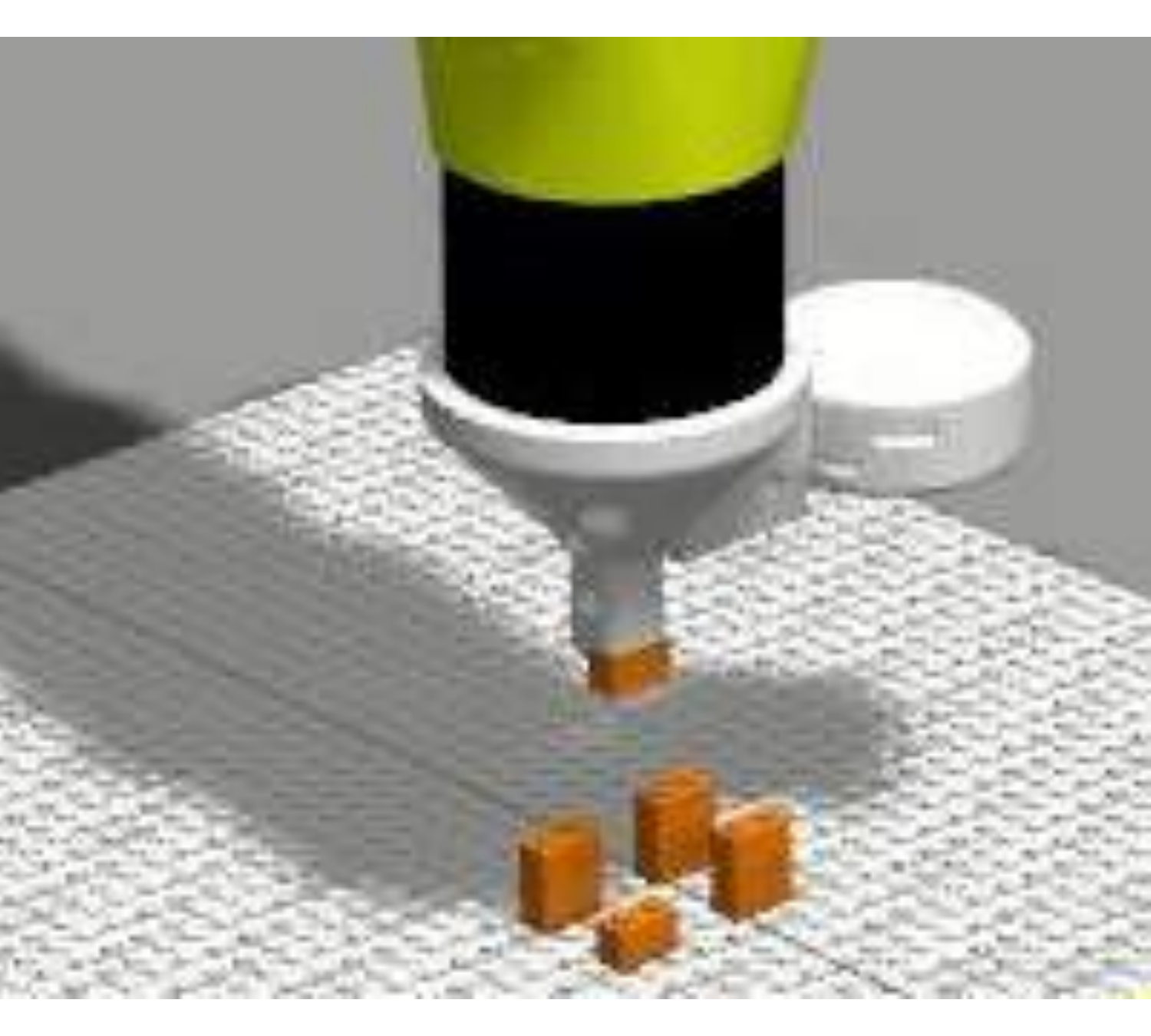}\label{fig:2}}\hfill
\subfigure[]{\includegraphics[width=0.16\linewidth]{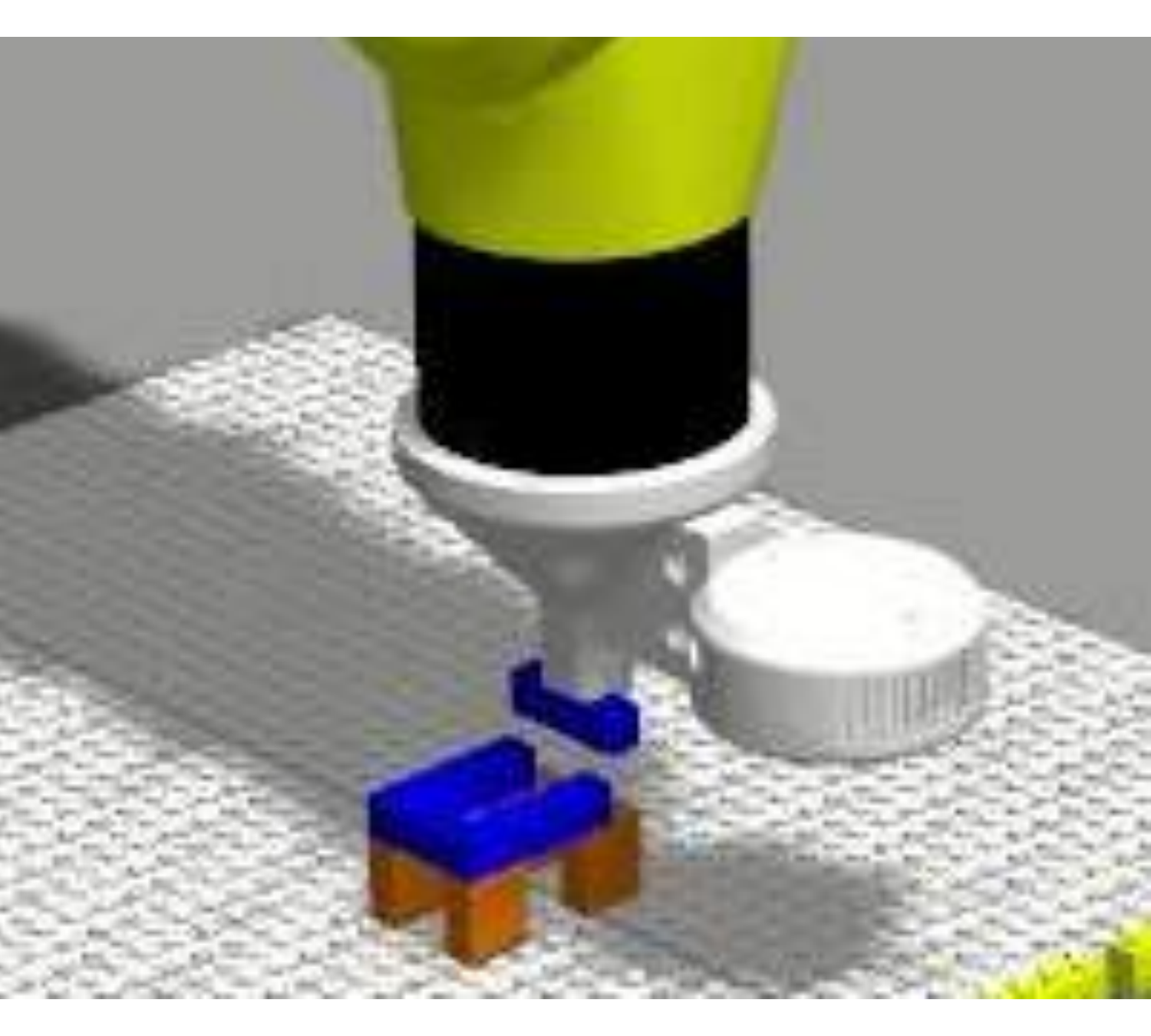}\label{fig:3}}\hfill
\subfigure[]{\includegraphics[width=0.16\linewidth]{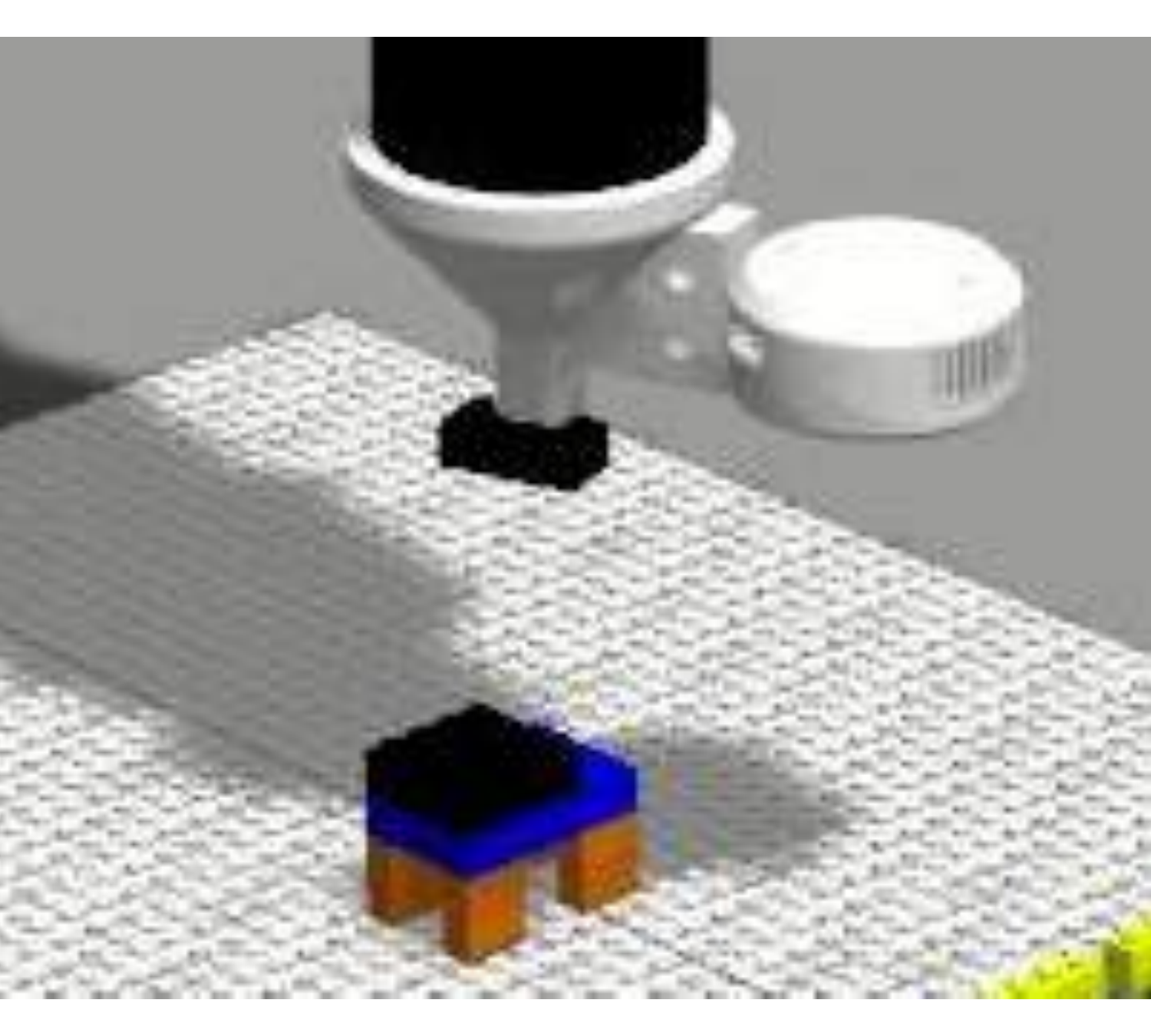}\label{fig:4}}\hfill
\subfigure[]{\includegraphics[width=0.16\linewidth]{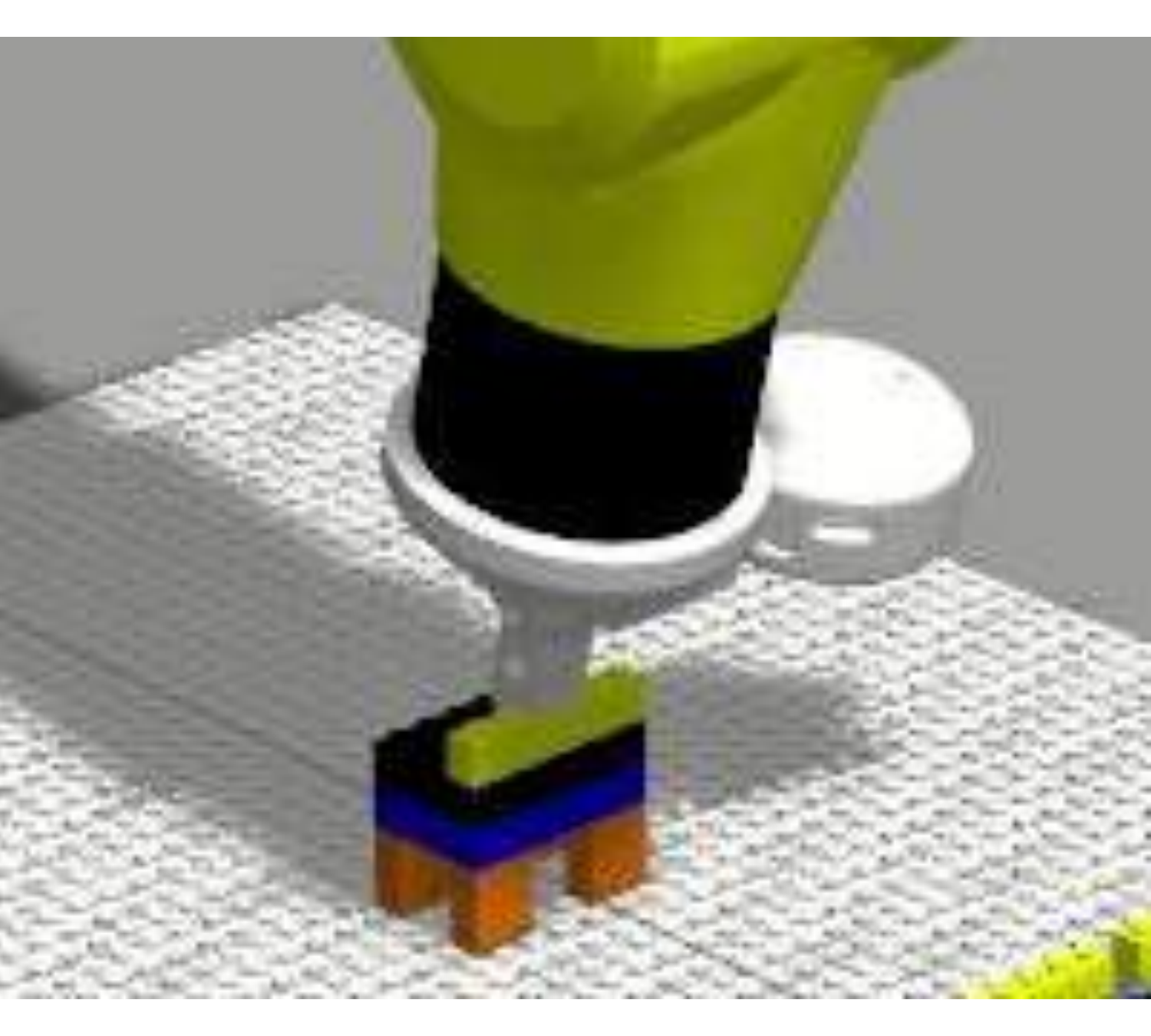}\label{fig:5}}\hfill
\subfigure[]{\includegraphics[width=0.16\linewidth]{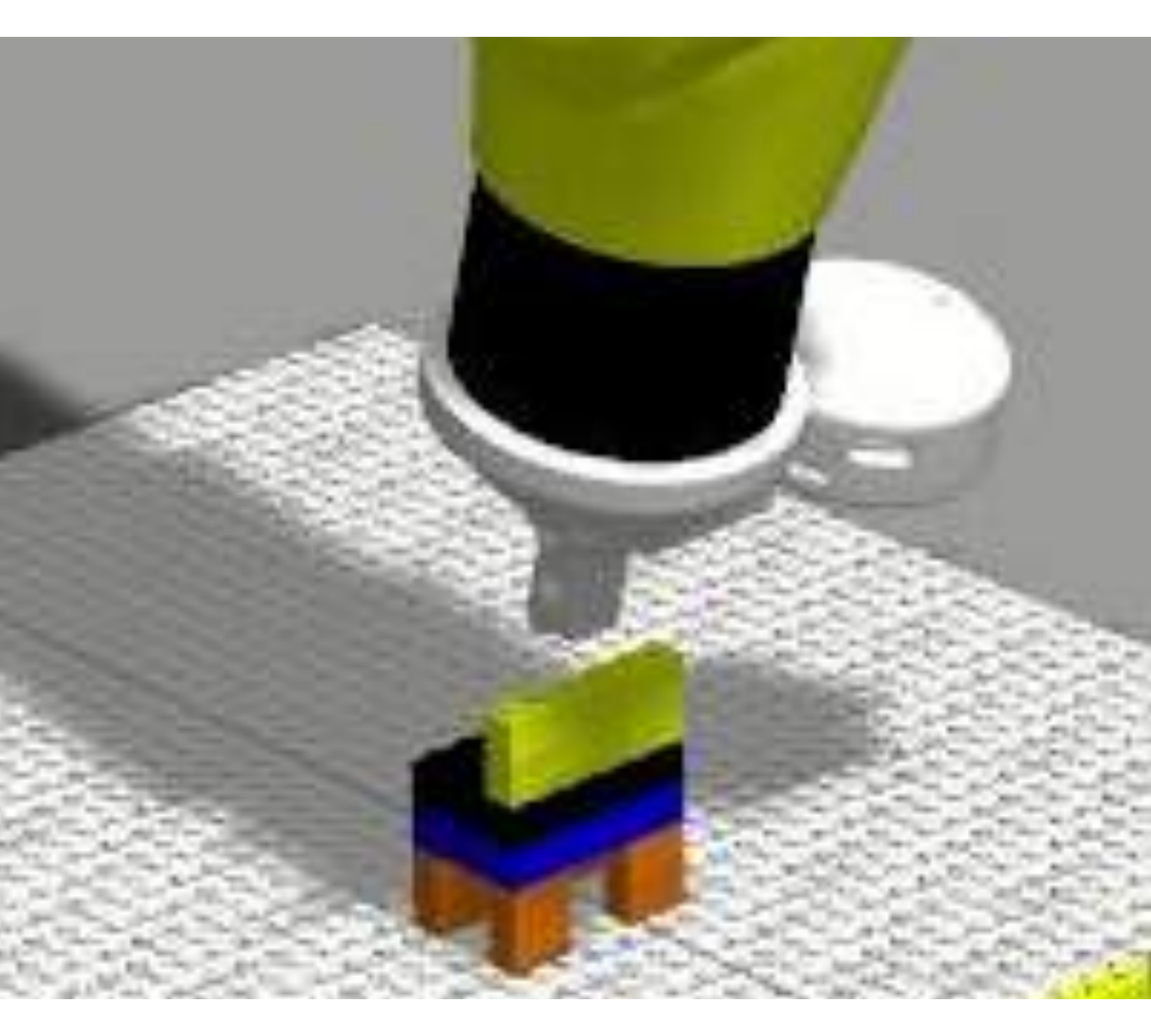}\label{fig:6}}
\\
\vspace{-10pt}
\subfigure[]{\includegraphics[width=0.16\linewidth]{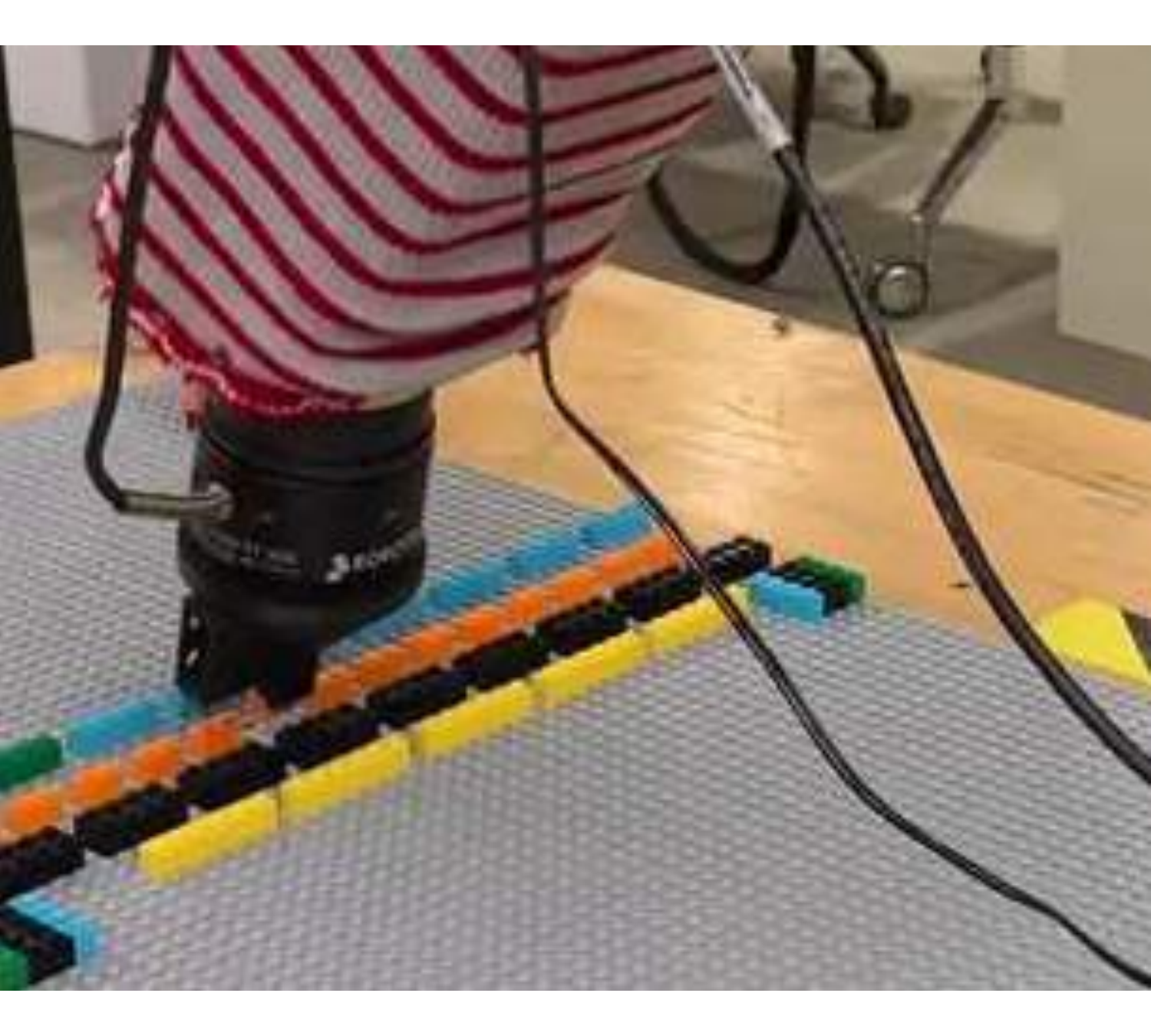}\label{fig:7}}\hfill
\subfigure[]{\includegraphics[width=0.16\linewidth]{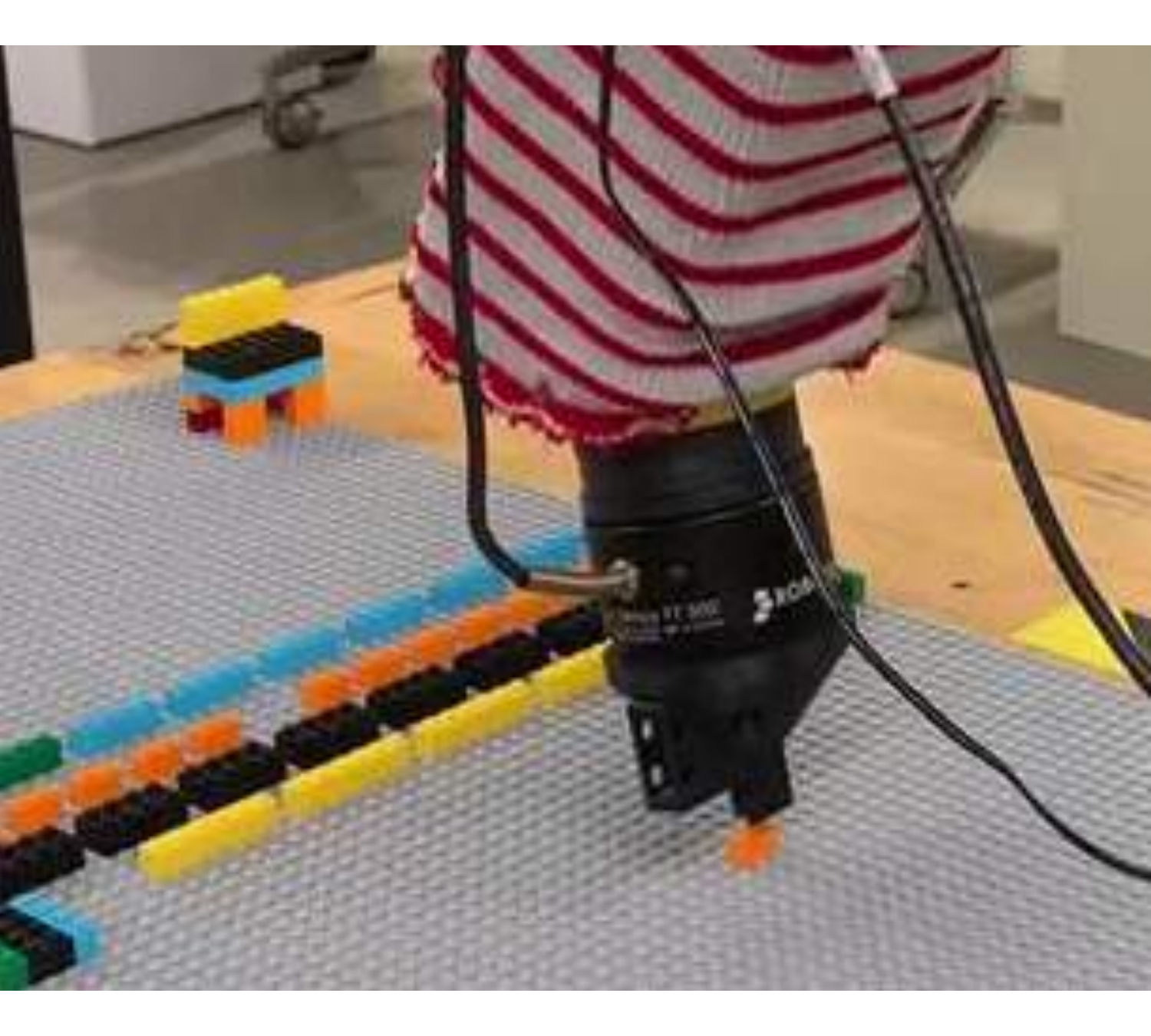}\label{fig:8}}\hfill
\subfigure[]{\includegraphics[width=0.16\linewidth]{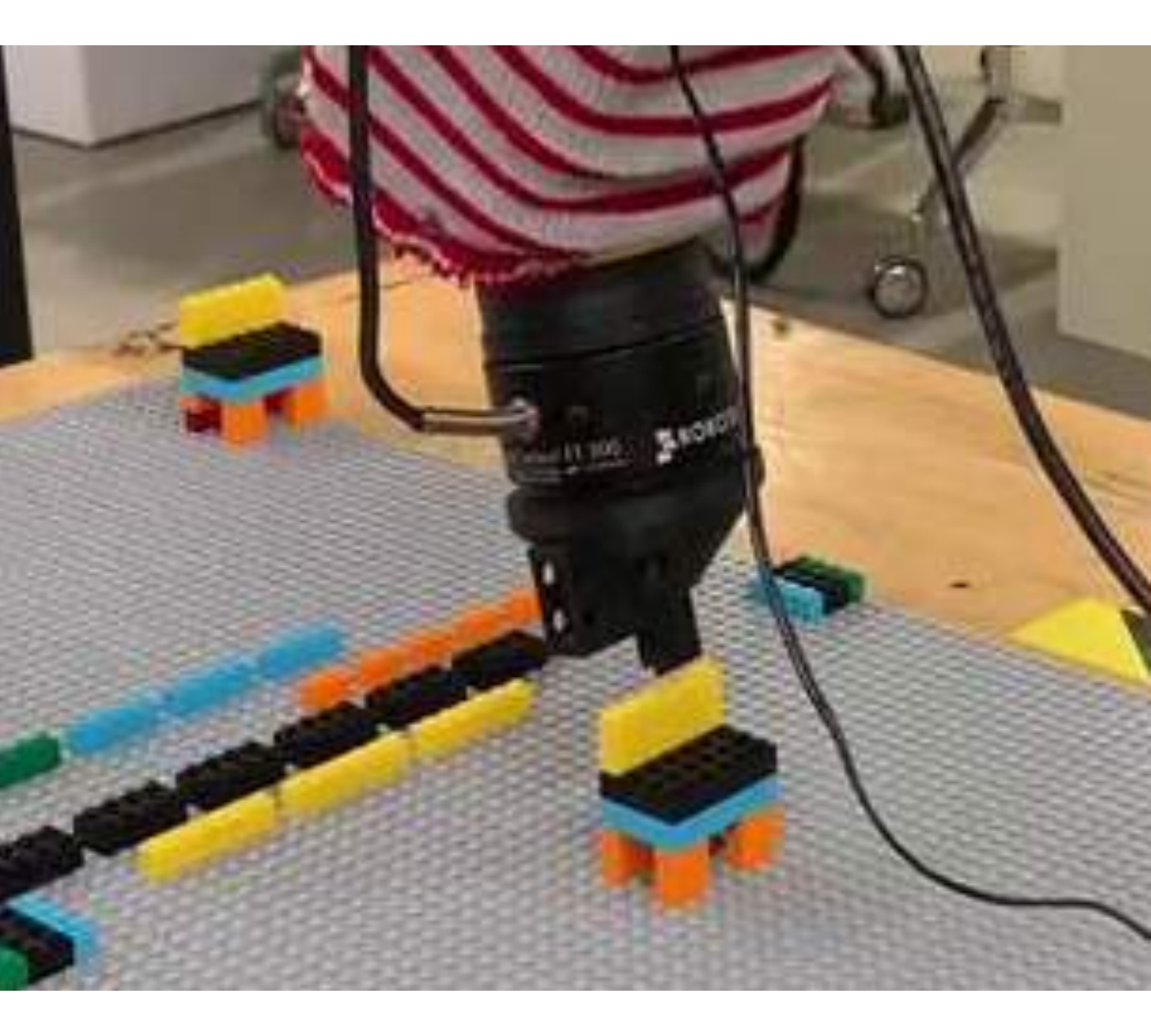}\label{fig:9}}\hfill
\subfigure[]{\includegraphics[width=0.16\linewidth]{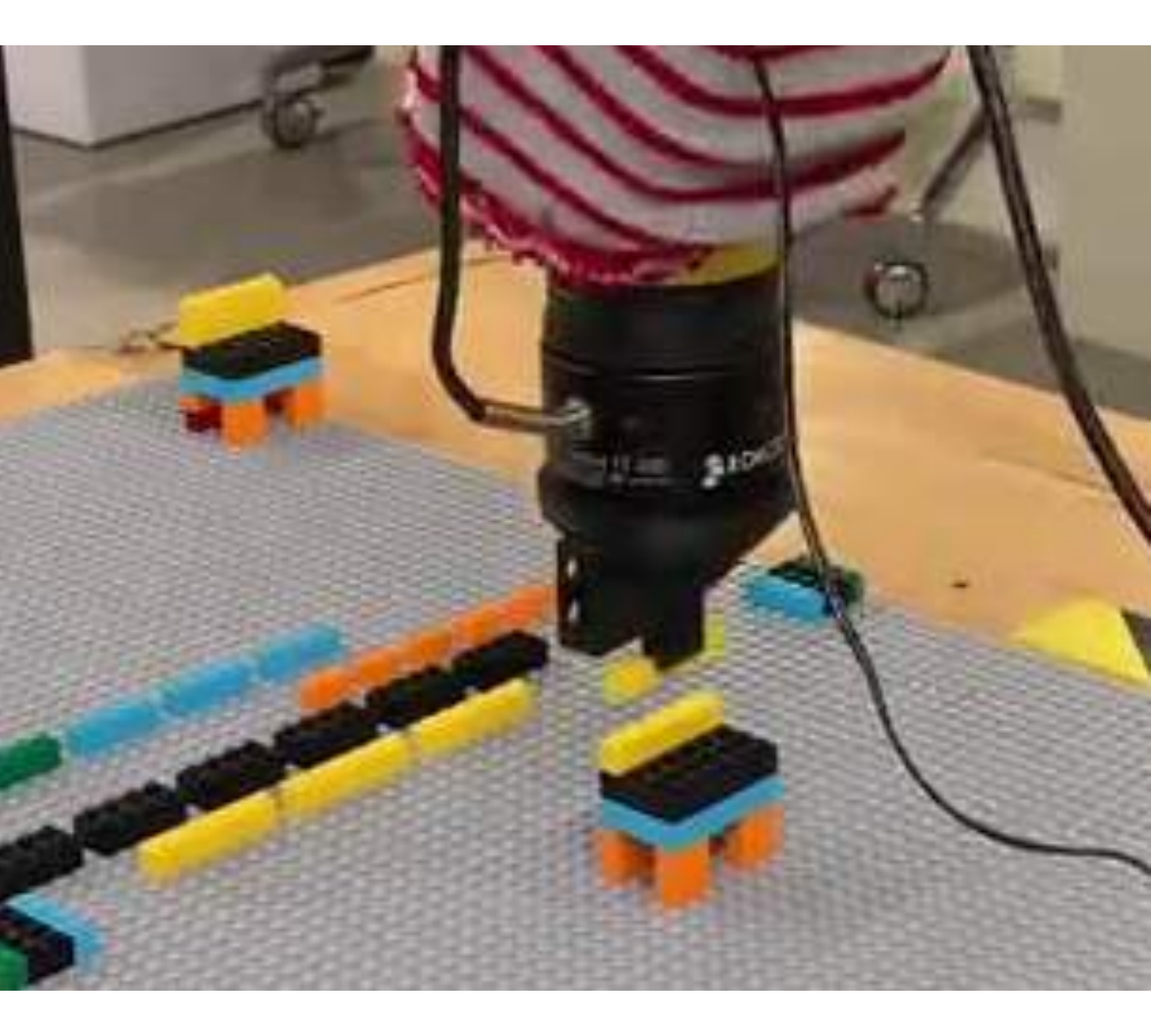}\label{fig:10}}\hfill
\subfigure[]{\includegraphics[width=0.16\linewidth]{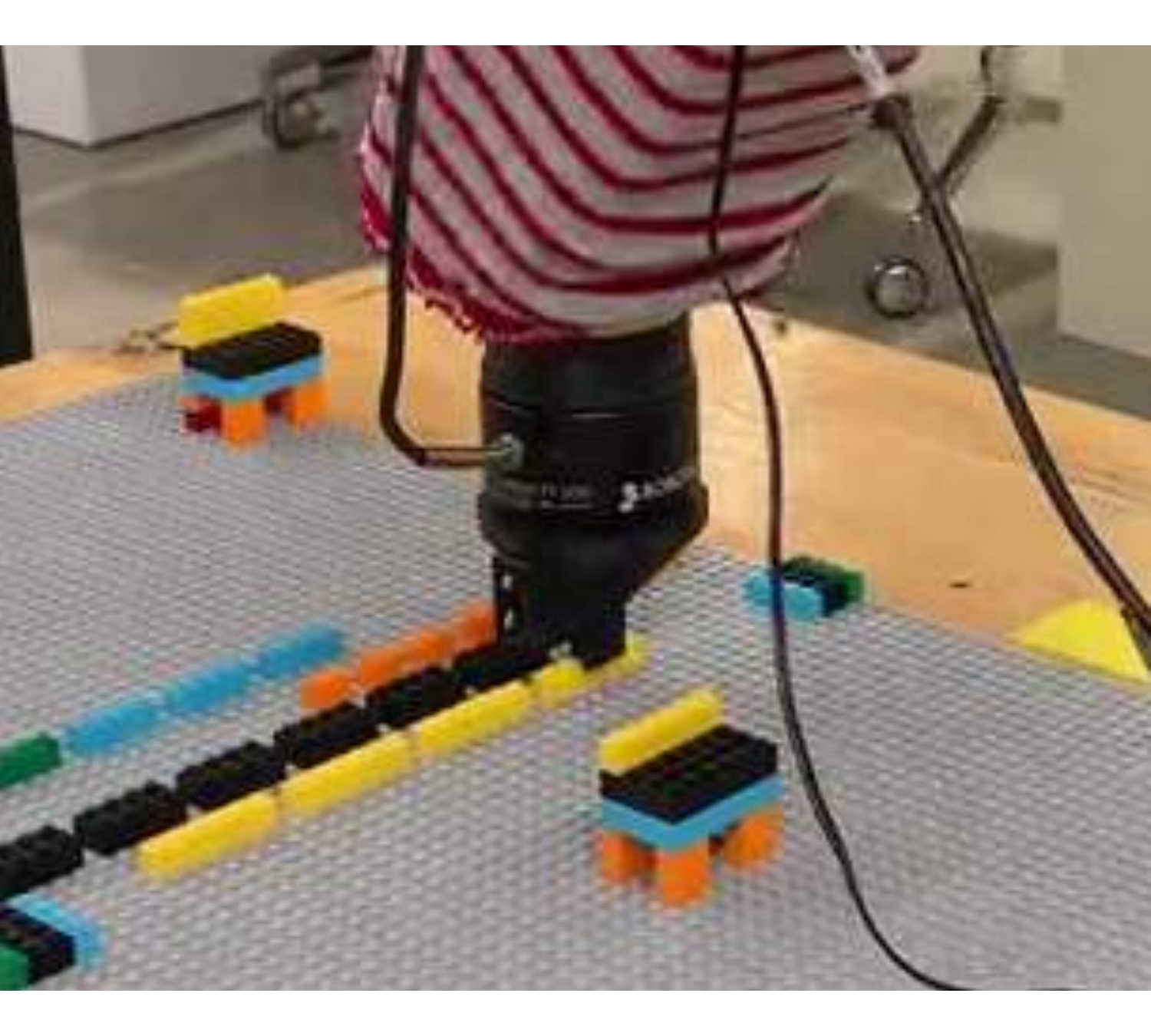}\label{fig:11}}\hfill
\subfigure[]{\includegraphics[width=0.16\linewidth]{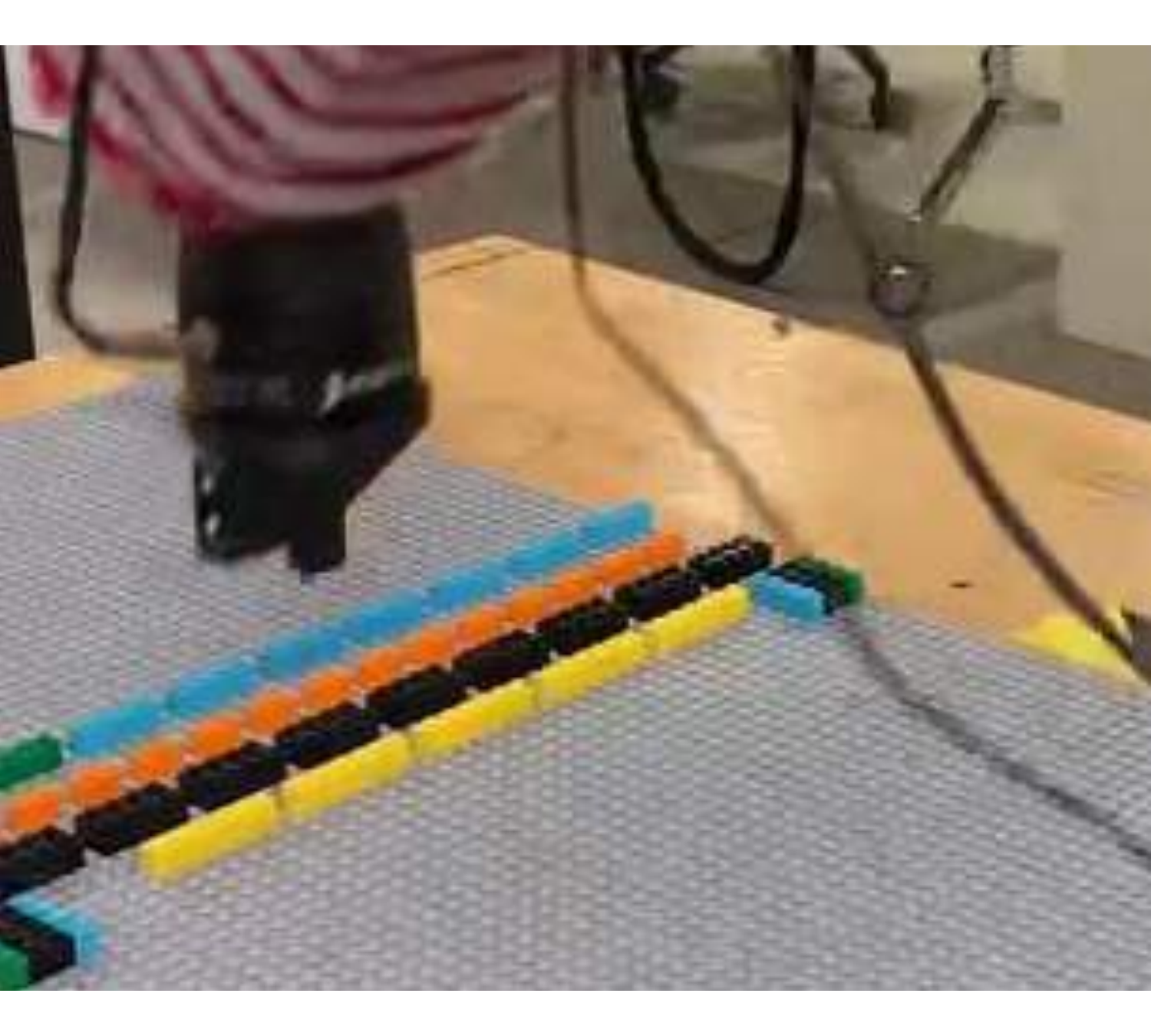}\label{fig:12}}
\vspace{-5pt}
    \caption{\footnotesize Simulation-aided robotic LEGO prototyping from human demonstration. Top: the human demonstrates building the LEGO prototype.
    Middle: simulation-aided LfD. Bottom: The robot executes the construction plan to assemble and disassemble the LEGO prototype. \label{fig:exp1}}
    \vspace{-15pt}
\end{figure*}

Besides the robustness of the proposed pipeline, we also conducted a comparison to demonstrate the efficiency of SaLfD for LEGO construction.
In particular, we compare handcrafting the construction plan, building the construction plan in simulation, and using SaLfD to learn the LEGO structure from demonstrations.
When handcrafting the construction plan, the user is given a built example LEGO.
The handcrafted plan does not need to build the LEGO at an identical global position.
Only the structure properties (\ie brick relative positions, orientations, and brick types) are needed.
In addition, the user does not need to generate the plan in a machine-readable format, which significantly reduces the human effort.
On the other hand, when generating the structure in simulation, the user can drag the bricks to the desired position.
The position, orientation, and brick type are given after the drag is completed.
However, the user is required to compile the plan into a machine-readable format, which is a JSON file.
For SaLfD, the human demonstrates and the pipeline generates the plan as shown in \cref{fig:lfd_pipeline}.
Each model is built two times, and
\cref{table:comparison_eff} shows the average time for each scenario.
The time for generating in simulation is significantly longer than others.
This is mainly because typing the tasks into machine-readable format is time-consuming.
It is possible to write a script to efficiently convert the format.
However, this would require the users to be equipped with programming knowledge.
On the other hand, handcrafting a LEGO construction plan is also time-consuming.
As the number of bricks increases, the time cost increases significantly.
In addition, when a structure has most of the bricks spread out, the time needed for each brick increases (\ie AI/RI vs Bridge/spiral).
This is because counting the brick position on the plate is time-consuming.
When bricks are close to each other or stacked straight up, the required effort is less.
It is obvious that SaLfD is more efficient as the time cost is significantly lower.
The time for SaLfD mainly includes the time for the human to demonstrate and the pipeline runtime.
Also, when demonstrating, the user only needs to place the desired brick at the desired position, which is more convenient and requires less effort.

\subsection{LEGO Construction from Human Demonstration}

In this section, we demonstrate robotic LEGO construction from human demonstration.

\paragraph{Assembly}
We provide text prompts and a single-view example image to the user as a guide for individual construction.
The users can then build their own structure.
This is mainly to avoid random construction and better evaluate the experiment.
The user is given a set of available bricks and builds the target structure on the assembling plate.
The SaLfD generates the robot-executable construction plan following the human assembly sequence.
The robot controller is implemented using \cite{jpc,jssa}.
The robot uses the available bricks on the storage plate, which is pre-defined, and assembles the target LEGO object in the assembling workspace.
\Cref{fig:prototypes} shows some example objects with the text prompts.
\Cref{fig:exp1} illustrates the assembly process.
The top row shows the human assembling a chair.
The SaLfD learns the task in the middle row.
\Cref{fig:7,fig:8,fig:9} shows the robot accomplishes the assembly of the LEGO chair.

\paragraph{Disassembly}
After the assembly is done, the robot then disassembles the structure and puts the bricks back into the storage space as shown in \cref{fig:10,fig:11,fig:12}.
Unlike the assembly, the user does not need to demonstrate the disassembly process.
The disassembly task is generated by reversing the assembly tasks.
In our experiment, the assembly and disassembly time increases approximately linearly with respect to the number of bricks of the LEGO structure.
Since \cite{jpc} allows the user to specify the task execution time, our current pipeline can assemble or disassemble a brick within $5$\si{\second}.
The full experiment video is attached in the supplementary.

\subsection{Discussion}
The proposed SaLfD is promising as illustrated by the experiment results.
However, there are limitations and future works.
First, the current framework only considers the scenario for single-arm assembly. 
In the future, we aim to extend it to a multi-agent system, where multiple robots collaborate and build more complex LEGO structures.
Second, the current pipeline is environment-specific, meaning that it is not robust to changes in lighting conditions, camera viewpoint, etc.
In the future, we aim to improve the system's robustness.
One approach could be to further minimize the sim-to-real gap (\eg camera field-of-view) so that virtual data can be used as an efficient data augmentation approach to alleviate the effort of data collection and improve the robustness. 

\section{Conclusion}
This paper studies learning LEGO construction from human demonstration.
It provides an easy and intuitive interface for the general public to interact with robots.
In particular, we present a simulation-aided LfD pipeline that is able to improve the learned LEGO construction plan for both assembly and disassembly.
The system is deployed to a FANUC LR-mate 200id/7L robot and the experiments demonstrate that it can effectively learn the tasks from human demonstrations.

\bibliographystyle{ifacconf}
\bibliography{ifacconf}      
\end{document}